\documentclass[10pt,a4paper]{article}

\usepackage[T1]{fontenc}
\usepackage[utf8]{inputenc}
\usepackage{times}
\usepackage[margin=2.5cm]{geometry}
\usepackage{amsmath,amssymb}
\usepackage{graphicx}
\usepackage{booktabs}
\usepackage{multirow}
\usepackage{makecell}
\usepackage{xcolor}
\usepackage[font=small,labelfont=bf]{caption}
\usepackage{subcaption}
\usepackage[numbers,sort&compress]{natbib}
\usepackage{microtype}
\usepackage{enumitem}
\usepackage{tikz}
\usetikzlibrary{arrows.meta,positioning,fit,calc}
\usepackage{url}      
\usepackage{seqsplit} 
\usepackage{needspace}
\usepackage[colorlinks=true,linkcolor=blue!50!black,citecolor=blue!50!black,urlcolor=blue!50!black]{hyperref}
\usepackage[capitalise,noabbrev]{cleveref}

\setlist{itemsep=1pt,topsep=3pt}

\newcommand{\AdvioFourShareFour}{61\,\%}

\newcommand{\ArkitHeightFive}{1.32\,m}
\newcommand{\ArkitHeightTwo}{1.43\,m}
\newcommand{\ArkitPitchFive}{$-1.2^\circ$}
\newcommand{\ArkitPitchTwo}{$+0.7^\circ$}

\newcommand{\BenchEgoGainAFour}{1.19}

\newcommand{\BenchEgoGainFour}{1.37}

\newcommand{\BenchEgoGainFourClusterCI}{[1.34, 1.40]}
\newcommand{\BenchEgoGainGoalsFour}{0.59}

\newcommand{\BenchEgoGainOne}{0.99}

\newcommand{\BenchEgoGainTwo}{1.22}

\newcommand{\BenchEgoGoalsADE}{0.59}

\newcommand{\BenchEgoGoalsFDE}{0.88}

\newcommand{\BenchEgoHeatCovFiftyFour}{0.58}
\newcommand{\BenchEgoHeatCovFiftyOne}{0.82}
\newcommand{\BenchEgoHeatCovFiftyTwo}{0.69}
\newcommand{\BenchEgoHeatCovNinetyFour}{0.93}
\newcommand{\BenchEgoHeatCovNinetyOne}{0.95}
\newcommand{\BenchEgoHeatCovNinetyTwo}{0.94}
\newcommand{\BenchEgoHeatNllFour}{2.94}

\newcommand{\BenchEgoNetAloneCovNinetyFour}{0.84}

\newcommand{\BenchEgoNetAloneNllFour}{2.86}

\newcommand{\BenchEgoTierAGainFour}{0.18}

\newcommand{\BenchSddGainAFour}{0.92}

\newcommand{\BenchSddGainFour}{1.51}

\newcommand{\BenchSddGainFourClusterCI}{[1.31, 1.70]}
\newcommand{\BenchSddGainGoalsFour}{0.37}

\newcommand{\BenchSddGainOne}{0.44}

\newcommand{\BenchSddGainTwo}{1.01}

\newcommand{\BenchSddGoalsADEpx}{13.54}

\newcommand{\BenchSddGoalsFDEpx}{19.71}

\newcommand{\BenchSddGoalsOneClusterCI}{[$-0.06$, $+0.08$]}
\newcommand{\BenchSddGoalsOneWindowCI}{[$+0.00$, $+0.06$]}
\newcommand{\BenchSddHeatArea}{22.8}
\newcommand{\BenchSddHeatCovFiftyFour}{0.65}
\newcommand{\BenchSddHeatCovFiftyOne}{0.88}
\newcommand{\BenchSddHeatCovFiftyTwo}{0.76}
\newcommand{\BenchSddHeatCovNinetyFour}{0.93}
\newcommand{\BenchSddHeatCovNinetyOne}{0.98}
\newcommand{\BenchSddHeatCovNinetyTwo}{0.96}

\newcommand{\BenchSddNetAloneArea}{12.3}

\newcommand{\BenchSddNetAloneCovNinetyFour}{0.86}

\newcommand{\BenchSddTierAGainFour}{0.59}

\newcommand{\BeyondTwentyEightHeld}{11\,\%}
\newcommand{\BlurZeroBands}{seven}
\newcommand{\ClipBootFourCI}{[$-0.73$, $-0.05$]}
\newcommand{\ClipBootOneCI}{[$-0.51$, $+0.14$]}
\newcommand{\ClipBootTwoCI}{[$-0.51$, $+0.11$]}
\newcommand{\ClipFiveArkitFour}{$-2.33$}

\newcommand{\ClipFiveArkitOne}{$-0.53$}

\newcommand{\ClipFiveArkitTwo}{$-1.21$}

\newcommand{\ClipFiveReferenceFour}{$+0.24$}

\newcommand{\ClipFiveReferenceOne}{$+0.42$}

\newcommand{\ClipFiveReferenceTwo}{$+0.46$}

\newcommand{\ClipTwoArkitFour}{$-1.08$}

\newcommand{\ClipTwoArkitOne}{$-0.70$}

\newcommand{\ClipTwoArkitTwo}{$-1.01$}

\newcommand{\ClipTwoReferenceFour}{$+0.10$}

\newcommand{\ClipTwoReferenceOne}{$-0.92$}

\newcommand{\ClipTwoReferenceTwo}{$-0.72$}

\newcommand{\CollapseNFixed}{803}
\newcommand{\CollapseNFollow}{1,107}
\newcommand{\CollapseShare}{31\,\%}
\newcommand{\DevClipsSigAll}{four}

\newcommand{\DevCovNinetyCVFour}{0.61}
\newcommand{\DevCovNinetyCVOne}{0.84}
\newcommand{\DevCovNinetyCVTwo}{0.71}
\newcommand{\DevCovNinetyFour}{0.73}
\newcommand{\DevCovNinetyOne}{0.89}
\newcommand{\DevCovNinetyTwo}{0.83}

\newcommand{\DevFarDiffOne}{$-0.44$}

\newcommand{\DevGainAFour}{1.03}
\newcommand{\DevGainAOne}{0.34}
\newcommand{\DevGainATwo}{0.72}
\newcommand{\DevGainFour}{1.21}
\newcommand{\DevGainFourCI}{[1.03, 1.39]}
\newcommand{\DevGainOne}{0.43}
\newcommand{\DevGainOneCI}{[0.38, 0.49]}

\newcommand{\DevGainSmoothOne}{0.71}

\newcommand{\DevGainTierACVOne}{0.09}

\newcommand{\DevGainTwo}{0.86}
\newcommand{\DevGainTwoCI}{[0.77, 0.96]}
\newcommand{\DevHeightMedian}{1.47\,m}

\newcommand{\DevMapEffectOne}{0.03}
\newcommand{\DevMapEffectOneCI}{[0.01, 0.04]}

\newcommand{\DevMidDiffOne}{$-0.37$}

\newcommand{\DevNNearFour}{48}
\newcommand{\DevNOne}{3,135}

\newcommand{\DevNearShareOne}{12\,\%}

\newcommand{\DevPeopleOne}{462}

\newcommand{\DriftArkitMaxAbs}{1.2\,m}
\newcommand{\DriftElevenArkitAbs}{0.48\,m}

\newcommand{\DriftSixArkitAbs}{1.23\,m}
\newcommand{\DriftSixBaro}{+0.33\,m}
\newcommand{\DriftSixGT}{$-$0.04\,m}
\newcommand{\EgoTestAgents}{9,828}
\newcommand{\EgoTestN}{9,828}

\newcommand{\EgoTestUses}{9}
\newcommand{\EgoTrainN}{25,381}
\newcommand{\EgoValN}{5,638}
\newcommand{\EscalatorFourGT}{$-$16.3\,m}
\newcommand{\EscalatorFourGTAbs}{16\,m}
\newcommand{\EscalatorOneGT}{+7.8\,m}
\newcommand{\EscalatorOneGTAbs}{7.8\,m}
\newcommand{\FiveNearShareArkit}{27\,\%}
\newcommand{\FiveNearShareReference}{68\,\%}
\newcommand{\FloorCVAdvioTwoOne}{56}

\newcommand{\FloorCVNOne}{69}

\newcommand{\FloorCVShareFour}{13.1\,\%}
\newcommand{\FloorCVShareOne}{3.7\,\%}

\newcommand{\FloorContribOne}{0.13}

\newcommand{\FloorElevenNearFixed}{32\,\%}
\newcommand{\FloorElevenNearFollow}{20\,\%}
\newcommand{\FloorFourFixedN}{28}
\newcommand{\FloorFourFixedNoTrack}{93\,\%}
\newcommand{\FloorFourFollowN}{729}
\newcommand{\FloorOneNearFixed}{0\,\%}
\newcommand{\FloorOneNearFollow}{19\,\%}

\newcommand{\FloorOursNOne}{9}

\newcommand{\FloorOursShareFour}{10.2\,\%}
\newcommand{\FloorOursShareOne}{0.5\,\%}

\newcommand{\FloorSixNearFixed}{7\,\%}
\newcommand{\FloorSixNearFollow}{20\,\%}
\newcommand{\FloorTwentyTwoNearFixed}{57\,\%}
\newcommand{\FloorTwentyTwoNearFollow}{7\,\%}
\newcommand{\FloorTwentyTwoNoTrackFollow}{0\,\%}

\newcommand{\FusionValEgoWOne}{1.77}
\newcommand{\FusionValEgoWZero}{2.49}

\newcommand{\FusionValSddWOne}{1.08}
\newcommand{\FusionValSddWZero}{1.37}

\newcommand{\HeldClipFiveDiffOne}{$+0.42$}
\newcommand{\HeldClipFiveDiffOneCI}{[$+0.18$, $+0.68$]}

\newcommand{\HeldClipsBetterFour}{four}
\newcommand{\HeldClipsBetterOne}{five}

\newcommand{\HeldClipsSigOne}{three}

\newcommand{\HeldCovEightyCVFour}{0.73}
\newcommand{\HeldCovEightyCVOne}{0.80}
\newcommand{\HeldCovEightyCVTwo}{0.75}
\newcommand{\HeldCovEightyFour}{0.58}
\newcommand{\HeldCovEightyOne}{0.75}
\newcommand{\HeldCovEightyTwo}{0.65}
\newcommand{\HeldCovFiftyCVFour}{0.66}
\newcommand{\HeldCovFiftyCVOne}{0.64}
\newcommand{\HeldCovFiftyCVTwo}{0.62}
\newcommand{\HeldCovFiftyFour}{0.33}
\newcommand{\HeldCovFiftyOne}{0.54}
\newcommand{\HeldCovFiftyTwo}{0.41}
\newcommand{\HeldCovNinetyCVFarFour}{0.34}
\newcommand{\HeldCovNinetyCVFour}{0.77}
\newcommand{\HeldCovNinetyCVOne}{0.86}
\newcommand{\HeldCovNinetyCVTwo}{0.80}
\newcommand{\HeldCovNinetyFarFour}{0.43}
\newcommand{\HeldCovNinetyFour}{0.76}
\newcommand{\HeldCovNinetyOne}{0.86}
\newcommand{\HeldCovNinetyTwo}{0.79}
\newcommand{\HeldFarDiffFour}{$-0.77$}
\newcommand{\HeldFarDiffFourCI}{[$-1.04$, $-0.46$]}
\newcommand{\HeldFarDiffOne}{$-0.42$}
\newcommand{\HeldFarDiffOneCI}{[$-0.63$, $-0.24$]}

\newcommand{\HeldGainAFour}{0.83}
\newcommand{\HeldGainAOne}{0.79}
\newcommand{\HeldGainATwo}{0.93}
\newcommand{\HeldGainFour}{0.47}
\newcommand{\HeldGainFourCI}{[0.11, 0.75]}
\newcommand{\HeldGainOne}{0.17}
\newcommand{\HeldGainOneCI}{[0.06, 0.27]}

\newcommand{\HeldGainTwo}{0.23}
\newcommand{\HeldGainTwoCI}{[0.04, 0.40]}

\newcommand{\HeldGoalsCVNegOne}{0.58}

\newcommand{\HeldMapDiffFour}{$+0.07$}

\newcommand{\HeldMapDiffOne}{$+0.06$}
\newcommand{\HeldMapDiffOneCI}{[$-0.02$, $+0.15$]}
\newcommand{\HeldMapDiffTwo}{$+0.10$}

\newcommand{\HeldMidDiffFour}{$+0.07$}
\newcommand{\HeldMidDiffFourCI}{[$-0.33$, $+0.40$]}
\newcommand{\HeldMidDiffOne}{$-0.41$}
\newcommand{\HeldMidDiffOneCI}{[$-0.58$, $-0.25$]}

\newcommand{\HeldMissNinetyRange}{14--24\,\%}
\newcommand{\HeldNFixed}{1,169}

\newcommand{\HeldNOne}{1,870}
\newcommand{\HeldNSens}{1,891}

\newcommand{\HeldNearDiffFour}{$-0.44$}
\newcommand{\HeldNearDiffFourCI}{[$-0.86$, $+0.19$]}
\newcommand{\HeldNearDiffOne}{$+0.01$}
\newcommand{\HeldNearDiffOneCI}{[$-0.13$, $+0.18$]}
\newcommand{\HeldNearDiffTwo}{$-0.06$}
\newcommand{\HeldNearDiffTwoCI}{[$-0.30$, $+0.28$]}

\newcommand{\HeldNearShareFourFive}{87\,\%}
\newcommand{\HeldNearShareOne}{56\,\%}

\newcommand{\HeldPeopleOne}{337}

\newcommand{\HeldSensFour}{0.47}

\newcommand{\HeldSensOne}{0.15}

\newcommand{\HeldSensTwo}{0.22}

\newcommand{\HeldSmoothFour}{0.32}
\newcommand{\HeldSmoothFourCI}{[$-$0.09, 0.63]}
\newcommand{\HeldSmoothOne}{0.16}
\newcommand{\HeldSmoothOneCI}{[0.04, 0.27]}
\newcommand{\HeldSmoothTwo}{0.13}
\newcommand{\HeldSmoothTwoCI}{[$-$0.08, 0.34]}

\newcommand{\HeldTierACVNegOne}{0.61}

\newcommand{\HiddenOverlayHeld}{9\,\%}
\newcommand{\InflationSixEnd}{87\,\%}

\newcommand{\LatCpuFps}{4.3}

\newcommand{\LatCpuTrack}{135}
\newcommand{\LatFrames}{900}
\newcommand{\LatGpuForecast}{15}
\newcommand{\LatGpuFps}{12.0}
\newcommand{\LatGpuMap}{14}

\newcommand{\LatGpuNoDrawFps}{20}
\newcommand{\LatGpuRender}{34}
\newcommand{\LatGpuTotal}{83}
\newcommand{\LatGpuTrack}{20}
\newcommand{\LooTwoGainOne}{0.05}

\newcommand{\MedianOne}{$-0.19$}
\newcommand{\MedianOneCI}{[$-0.29$, $-0.09$]}

\newcommand{\MirrorBugEgoValMax}{1.801}
\newcommand{\MirrorBugEgoValMin}{1.772}
\newcommand{\MirrorBugSddValMax}{1.099}
\newcommand{\MirrorBugSddValMin}{1.077}
\newcommand{\MirrorBugSelectMax}{1.358}
\newcommand{\MirrorBugSelectMin}{1.338}

\newcommand{\MirrorFixEgoVal}{1.793}
\newcommand{\MirrorFixSddVal}{1.103}
\newcommand{\MirrorFixSelect}{1.343}

\newcommand{\MixCVGainFour}{0.94}
\newcommand{\MixCVGainOne}{0.30}
\newcommand{\MixCVGainTwo}{0.50}

\newcommand{\MixFarOne}{$-0.28$}
\newcommand{\MixFarOneCI}{[$-0.46$, $-0.12$]}
\newcommand{\MixFarTwo}{$-0.29$}
\newcommand{\MixFarTwoCI}{[$-0.52$, $-0.06$]}
\newcommand{\MixFour}{$+0.05$}
\newcommand{\MixFourCI}{[$-0.10$, $+0.21$]}

\newcommand{\MixMidOne}{$-0.42$}
\newcommand{\MixMidOneCI}{[$-0.55$, $-0.28$]}
\newcommand{\MixMidTwo}{$-0.30$}
\newcommand{\MixMidTwoCI}{[$-0.50$, $-0.10$]}

\newcommand{\MixNearOne}{$+0.02$}
\newcommand{\MixNearOneCI}{[$-0.06$, $+0.10$]}
\newcommand{\MixNearTwo}{$+0.15$}
\newcommand{\MixNearTwoCI}{[$+0.01$, $+0.33$]}
\newcommand{\MixOne}{$-0.14$}
\newcommand{\MixOneCI}{[$-0.20$, $-0.07$]}
\newcommand{\MixOneClipCI}{[$-0.37$, $+0.03$]}
\newcommand{\MixOursGainFour}{0.43}
\newcommand{\MixOursGainOne}{0.27}
\newcommand{\MixOursGainTwo}{0.31}
\newcommand{\MixTwo}{$-0.04$}
\newcommand{\MixTwoCI}{[$-0.15$, $+0.07$]}

\newcommand{\NearAdvioFourShareFour}{95\,\%}
\newcommand{\NearAdvioFourShareOne}{68\,\%}
\newcommand{\NearAdvioFourShareTwo}{78\,\%}

\newcommand{\NoFloorOne}{$-0.04$}
\newcommand{\NoFloorOneCI}{[$-0.14$, $+0.06$]}

\newcommand{\OursMixWeightRange}{13--41\,\%}
\newcommand{\OursMixWidthMax}{0.5\,m}
\newcommand{\OursVsMixCVFour}{$+0.48$}
\newcommand{\OursVsMixCVFourCI}{[$+0.19$, $+0.80$]}
\newcommand{\OursVsMixCVFourClipCI}{[$-0.11$, $+0.72$]}
\newcommand{\OursVsMixCVOne}{$+0.13$}
\newcommand{\OursVsMixCVOneCI}{[$+0.00$, $+0.26$]}
\newcommand{\OursVsMixCVOneClipCI}{[$-0.37$, $+0.44$]}

\newcommand{\OutsideGridFour}{7.4\,\%}

\newcommand{\PoseArkitFour}{$-0.64$}
\newcommand{\PoseArkitFourCI}{[$-0.93$, $-0.32$]}
\newcommand{\PoseArkitOne}{$-0.17$}
\newcommand{\PoseArkitOneCI}{[$-0.30$, $-0.04$]}

\newcommand{\PoseRefShareHeld}{69\,\%}
\newcommand{\PoseReferenceFour}{$-0.39$}
\newcommand{\PoseReferenceFourCI}{[$-0.79$, $+0.19$]}
\newcommand{\PoseReferenceOne}{$-0.18$}
\newcommand{\PoseReferenceOneCI}{[$-0.31$, $-0.03$]}

\newcommand{\ProtocolCommit}{db73188}
\newcommand{\ProtocolSigmaFarFour}{2.1\,m}

\newcommand{\RecalGridAgree}{97\,\%}
\newcommand{\RecalGridMeanDiff}{0.0004}

\newcommand{\RecalMismatchText}{none of}

\newcommand{\RecalRecalVsRefitFour}{$+0.07$}
\newcommand{\RecalRecalVsRefitFourCI}{[$-0.21$, $+0.41$]}

\newcommand{\RecalRecalVsRefitOne}{$-0.18$}
\newcommand{\RecalRecalVsRefitOneCI}{[$-0.28$, $-0.07$]}
\newcommand{\RecalRecalVsRefitOneClipCI}{[$-0.55$, $+0.04$]}
\newcommand{\RecalRecalVsRefitTwo}{$-0.10$}
\newcommand{\RecalRecalVsRefitTwoCI}{[$-0.26$, $+0.07$]}

\newcommand{\RecalSigmaFarFour}{5.7\,m}
\newcommand{\RefPeopleHeightFive}{1.87\,m}
\newcommand{\RefPeopleHeightTwo}{0.83\,m}
\newcommand{\RideForecastShareFour}{97\,\%}

\newcommand{\RideLagFour}{0.7\,m}

\newcommand{\RideScaleFour}{52\,\%}

\newcommand{\RideShareFour}{81\,\%}

\newcommand{\RideSpeedFour}{0.36}

\newcommand{\SddTestN}{2,829}
\newcommand{\SddTestUses}{10}
\newcommand{\SddTrainOnCore}{1.9\,\%}
\newcommand{\SddTrainOnStructure}{6.0\,\%}
\newcommand{\SddValN}{849}
\newcommand{\SddValScenes}{deathCircle\_4, hyang\_9, nexus\_9, gates\_8, bookstore\_3}

\newcommand{\SeedsEgoRangeMax}{0.05}

\newcommand{\SeedsSddRangeMax}{0.04}

\newcommand{\SignFlipFour}{0.09}
\newcommand{\SignFlipOne}{0.15}
\newcommand{\SignFlipTwo}{0.10}
\newcommand{\SlopeTwentyThreeBaro}{$-$4.23\,m}
\newcommand{\SlopeTwentyTwoArkit}{$-$1.79\,m}
\newcommand{\SlopeTwentyTwoBaro}{$-$1.54\,m}
\newcommand{\SmoothGridAgree}{94\,\%}
\newcommand{\SmoothRefitFour}{$+0.22$}
\newcommand{\SmoothRefitFourCI}{[$-0.09$, $+0.59$]}
\newcommand{\SmoothRefitOne}{$-0.02$}
\newcommand{\SmoothRefitOneCI}{[$-0.14$, $+0.10$]}
\newcommand{\SmoothRefitTwo}{$+0.17$}
\newcommand{\SmoothRefitTwoCI}{[$-0.04$, $+0.38$]}
\newcommand{\SynthDetections}{55}
\newcommand{\SynthFixedDownPlaced}{0}
\newcommand{\SynthFixedUpErr}{5.2\,m}
\newcommand{\SynthFollowDownErr}{0.09\,m}
\newcommand{\SynthFollowUpErr}{0.07\,m}
\newcommand{\TrainWindowsEgo}{25,381}
\newcommand{\TrainWindowsSdd}{7,644}
\newcommand{\TwoFarShareArkit}{19\,\%}
\newcommand{\TwoFarShareReference}{72\,\%}

\newcommand{\TwoFiveArkitNOne}{1,596}

\newcommand{\TwoFiveArkitOne}{$-0.66$}
\newcommand{\TwoFiveArkitOneCI}{[$-0.77$, $-0.56$]}

\newcommand{\TwoFiveReferenceNOne}{568}

\newcommand{\TwoFiveReferenceOne}{$-0.21$}

\title{\bfseries Probabilistic Pedestrian Forecasts from a Handheld Phone:\\
World-Frame Heat Maps, Visual-Inertial Height Drift,\\
and Evaluation without Ground Truth}

\author{Danial Safaei\\[2pt]
WMG, University of Warwick, Coventry, United Kingdom\\[2pt]
\href{mailto:danial.safaei@warwick.ac.uk}{\texttt{danial.safaei@warwick.ac.uk}}}

\date{}

\begin{document}
\maketitle

\begin{abstract}
A pedestrian with a phone could be shown where nearby people will be in the
next few seconds, if the forecast stays on the ground while the phone moves,
is calibrated, and needs only a monocular camera and visual-inertial odometry
(VIO). We build and evaluate such a system. People are detected, lifted onto
the floor by ray--plane intersection, tracked in a gravity-aligned metric
frame, and forecast as per-step probability maps by a small U-Net trained with
a negative log-likelihood (NLL) loss on bird's-eye (SDD) and first-person
(EgoTraj-Bench) trajectories. On handheld ADVIO recordings, vertical VIO drift
and the user's own changes of level silently rescale monocular ground
positions (by \InflationSixEnd{} within 90\,s on one clip; on another, all
tracks are lost for the last \CollapseShare{} of the clip); keeping the
camera's height above the floor constant under a low-pass-filtered altitude
avoids this, though it lags on escalators. On the SDD and EgoTraj-Bench test
splits, the final forecaster lowers the NLL at 4.8\,s by \BenchSddGainFour{}
and \BenchEgoGainFour{}~nats relative to a fitted constant-velocity Gaussian.
Lacking ground truth for people in handheld video, we score forecasts against
the tracker's own later raw measurements. In an internally pre-registered
evaluation on seven held-out clips, the final forecaster's NLL is lower than
the benchmark-fitted baseline's at 1.2, 2.4 and 4.8\,s (by \HeldGainOne{}, \HeldGainTwo{} and
\HeldGainFour{}~nats; 95\,\% intervals over people exclude zero), but by less
than half as much as on the development clips. Exploratory analyses cut both
ways: resampling clips instead of people widens the intervals to include zero
at 1.2 and 2.4\,s, and once both forecasters are recalibrated on the
development clips the network is significantly better only at 1.2\,s; but two
clips run with ADVIO's reference poses favour the network much more when re-run
with the phone's own poses. We discuss what such self-consistency scores can
and cannot show.
\end{abstract}

\section{Introduction}
\label{sec:intro}

A pedestrian who walks with a phone in front of them, for instance a blind or
low-vision user of a navigation aid, or anyone following augmented-reality
directions, would benefit from knowing where the people around them will be
in the next few seconds. Three properties make such a forecast useful. It
must be \emph{metric and attached to the world}, so that it stays on the
ground while the phone translates, pans and tilts. It must be
\emph{probabilistic and calibrated}: if the 90\,\% region is used to warn the
user or to plan a path, it should contain the person about 90\,\% of the
time. And it must come from what a phone has: a monocular camera and the
pose from its visual-inertial odometry (VIO), with no depth sensor on most
models and no fixed camera overlooking the scene.

Trajectory forecasting research mostly works with bird's-eye trajectories from
fixed or drone cameras~\citep{robicquet2016learning} and reports best-of-$K$
displacement errors~\citep{gupta2018social,mangalam2021goals}, which reward
spread-out forecasts~\citep{thiede2019analyzing} and say nothing about
calibration. First-person benchmarks record from robots or render from
simulators~\citep{wang2024tbd,liu2026egotraj,stoler2023t2fpv}; assistive systems
that use a handheld phone track people but do not forecast them~\citep{kuribayashi2021linechaser},
and head-mounted ones rely on stereo or offline SLAM~\citep{haghighi2024headsup,luan2025unified}.
Finally, there is no public handheld-phone recording of pedestrians with
ground-truth trajectories, so it is unclear how to measure such a forecaster on
real video at all.

This paper builds and evaluates such a system end to end
(\cref{fig:system}). People are detected, lifted onto the floor by intersecting
pixel rays with a floor plane, tracked in a gravity-aligned metric world frame,
and forecast as per-step probability maps over the ground, either by a
training-free scene-and-physics prior or by a small U-Net trained with a
(slightly smoothed) negative log-likelihood loss on top-down (SDD) and real
first-person (EgoTraj-Bench) trajectories. The forecasts are re-projected into each new frame
with the current pose. We report four findings.

\begin{enumerate}[leftmargin=*]
\item \textbf{Vertical VIO drift breaks monocular lifting, and a simple rule
avoids it.} On ADVIO's handheld recordings~\citep{cortes2018advio}, the
phone's reported height drifts by \DriftArkitMaxAbs{} within 90\,s while
ADVIO's reference poses stay level, and on streets and escalators the phone's
true height changes by metres. Holding the floor plane fixed in the world scales
every lifted distance by $(h+\Delta z)/h$; on one clip every track is lost for
the last \CollapseShare{} of the clip. Keeping the camera's height above the
floor constant under a low-pass-filtered altitude restores tracking on these
recordings and removes the rescaling in a synthetic test, although it lags
behind sustained climbs and descents such as escalator rides
(\cref{sec:floor,sec:exp_drift,sec:discussion}).
\item \textbf{A heat-map network trained with negative log-likelihood beats a
physics prior.} On the SDD TrajNet test split and the EgoTraj-Bench test split
the final forecaster, the network's own heat map (chosen on validation data
over a mixture with the physics rollout) with a small constant-velocity
component, lowers the negative log-likelihood of the true position relative to
a constant-velocity Gaussian baseline by \BenchSddGainFour{} and
\BenchEgoGainFour{}~nats at 4.8\,s (Tier A: \BenchSddTierAGainFour{} and
\BenchEgoTierAGainFour{}~nats), with 90\,\% regions smaller than those of the
CV baseline and of Tier A that contain the truth \BenchSddHeatCovNinetyFour{}
and \BenchEgoHeatCovNinetyFour{} of the time at 4.8\,s, slightly more often
than nominal (\cref{sec:exp_bench}).
\item \textbf{On real handheld video, part of the gain carries over in an
internally pre-registered test, but it is fragile.} We score forecasts against
the tracker's own later raw measurements, with forecast and baseline widened
by the same measurement noise. On seven ADVIO clips that were set aside until
the method and the analysis were frozen and committed, the final forecaster's
NLL is lower than that of the constant-velocity baseline, whose widths come
from benchmark training data, by \HeldGainOne{}, \HeldGainTwo{} and
\HeldGainFour{}~nats at 1.2, 2.4 and 4.8\,s, with 95\,\% intervals over people
that exclude zero. That is less than half of the gain on the development
clips; the shortfall lies in the nearest distance band, which one escalator
clip with known geometric problems dominates (\cref{sec:exp_heldout}). The
final forecaster's 90\,\% regions contain the later measurement only
\HeldCovNinetyOne{}, \HeldCovNinetyTwo{} and \HeldCovNinetyFour{} of the time.
Exploratory analyses show how much the comparison depends on the clips, on the
baseline and on the poses: with clips as the unit of resampling only the
4.8\,s difference excludes zero, and even it rests on four of seven clips; a
constant-velocity baseline refitted on the development clips as a mixture of
two Gaussians beats the final forecaster as configured, and when both are
recalibrated the network keeps a significant advantage (over people) only at
1.2\,s; but two clips that the protocol evaluated with ADVIO's offline
reference poses favour the network much more when re-converted with the
phone's own poses (\cref{sec:exp_explore}).
\item \textbf{Limits.} Best-of-20 errors remain far from the state of the
art; self-consistency scores measure where the tracker will see a person, not
where the person is; and the pipeline processes \LatGpuFps{} frames per
second on a workstation GPU, below the 30 of our converted clips
(\cref{sec:exp_runtime,sec:discussion}).
\end{enumerate}

The code, the protocol and every result file needed to regenerate the tables
and figures are available from the author on request.

\section{Related work}
\label{sec:related}

\paragraph{Human trajectory forecasting.}
Classical models treat walking as driven by goals and by repulsion from
obstacles and other people~\citep{helbing1995social}. Learned forecasters model
interactions with recurrent networks and pooling~\citep{alahi2016social,gupta2018social},
attention over the scene and the neighbours~\citep{sadeghian2019sophie,yuan2021agentformer},
graph-structured generative models~\citep{ivanovic2019trajectron,salzmann2020trajectron},
endpoint- or goal-conditioned generation~\citep{mangalam2020not,mangalam2021goals,chiara2022goal}
and diffusion~\citep{gu2022stochastic,mao2023leapfrog}. Our learned forecaster is
closest to Y-Net~\citep{mangalam2021goals}, which predicts per-pixel goal and
trajectory heat maps with a U-Net~\citep{ronneberger2015unet} on a bird's-eye
scene image. We keep the heat-map output but train it in metres, in a
heading-aligned crop, as a normalised distribution with a negative
log-likelihood loss, and on first-person noisy histories as well as top-down
ones. A constant-velocity model is a strong baseline that several learned
forecasters fail to beat~\citep{scholler2020what}; we use a
constant-velocity Gaussian with widths fitted to training data as the
reference throughout. Most of this work is
evaluated on fixed overhead or drone views such as the Stanford Drone
Dataset~\citep{robicquet2016learning} in the TrajNet split~\citep{sadeghian2018trajnet},
or on interaction benchmarks such as TrajNet++~\citep{kothari2022human}.

\paragraph{Evaluating probabilistic forecasts.}
Best-of-$K$ displacement errors~\citep{gupta2018social} reward forecasts that
are spread out: the variety loss behind them converges to the square root of
the true density rather than the density~\citep{thiede2019analyzing}, which is why
\citet{thiede2019analyzing} recommend reporting the likelihood of the ground truth
as well. Trajectron and Trajectron++ report a kernel-density negative
log-likelihood~\citep{ivanovic2019trajectron,salzmann2020trajectron}. The
logarithmic score is strictly proper~\citep{gneiting2007strictly}, and good
probabilistic forecasts maximise sharpness subject to calibration~\citep{gneiting2007probabilistic};
we therefore report negative log-likelihood together with the coverage and
area of highest-density regions~\citep{hyndman1996computing}. Recalibration~\citep{kuleshov2018accurate} and conformal
prediction~\citep{lindemann2023safe,dixit2023adaptive} can make the regions of
any forecaster calibrated, the latter with finite-sample coverage guarantees;
we measure coverage directly and use only simple parametric refits (one width
or a two-component mixture), in an exploratory analysis.

\paragraph{First-person and moving-camera forecasting.}
Forecasting from video rather than from coordinates alone can exploit the
appearance of people and scenes~\citep{liang2019peeking}. Forecasting from a
moving, first-person view has been studied in image
coordinates for chest-mounted cameras~\citep{yagi2018future} and wearable
cameras with inertial cues~\citep{qiu2021indoor}, and relative to a moving
vehicle~\citep{neumann2021pedestrian,rasouli2019pie,rasouli2017they}. Robot
datasets provide egocentric sensing with metric labels, such as
JRDB~\citep{martinmartin2023jrdb} and TBD~\citep{wang2024tbd}. T2FPV renders
first-person views of the ETH/UCY pedestrians to expose detection, tracking and
field-of-view errors~\citep{stoler2023t2fpv}, and EgoTraj-Bench pairs real noisy
first-person histories, obtained with a detector and BoT-SORT~\citep{aharon2022botsort}
on the first-person video of the TBD platform, with clean metric
futures~\citep{liu2026egotraj}.
RealTraj~\citep{fujii2026realtraj} makes forecasting robust to perception errors
by forecasting from past detections instead of tracks. Head-mounted systems
for blind users forecast pedestrians from a stereo camera with its visual-inertial
odometry~\citep{haghighi2024headsup}, or jointly localise and forecast from a
monocular camera with an offline SLAM ego-pose~\citep{luan2025unified}. To our
knowledge, none of these evaluates density forecasts, or their calibration,
from a handheld phone.

\paragraph{Phones and assistive navigation.}
LineChaser guides blind users in queues with a handheld iPhone, localising the
user with ARKit visual-inertial odometry and placing detected people in a map,
but does not forecast them~\citep{kuribayashi2021linechaser}. A suitcase-shaped
guide with RGB-D cameras and LiDAR tracks pedestrians and extrapolates them with
constant velocity four seconds ahead to issue collision alerts~\citep{kayukawa2020guiding}.
A chest-mounted phone LiDAR has been used to estimate time to impact~\citep{rawat2025impactalert}.
To our knowledge, no prior work forecasts other pedestrians as probability
distributions in a world frame from a handheld phone's camera and
visual-inertial pose and measures their calibration on real handheld video.

\paragraph{Monocular metric geometry, tracking and drift.}
Without depth, the ground position of a person follows from intersecting the
ray through their feet with the ground plane, which requires the camera's height
and attitude. The known height of people constrains both~\citep{hoiem2006putting,hoiem2008putting};
we use this ``people as the ruler'' idea to estimate the camera height.
Tracking people on the ground plane rather than in the image makes the motion
model independent of camera motion~\citep{yi2024ucmctrack}; we track with
Kalman filters~\citep{kalman1960new} in a gravity-aligned world frame and use
a Rauch--Tung--Striebel smoother~\citep{rauch1965maximum} for evaluation. Image-space
multi-object trackers~\citep{zhang2022bytetrack,aharon2022botsort} are
complementary. The ADVIO benchmark~\citep{cortes2018advio}, which we use, compared
the visual-inertial odometry of ARKit, ARCore and Tango with ground truth and
found that ARKit ``had severe problems with drifting in the outdoor
sequences'' and that drift was worse in multi-floor cases. We show that the
\emph{vertical} component of such drift silently rescales every monocular
ground-plane position, and how to avoid it.

\section{Method}
\label{sec:method}

\begin{figure}[t]
\centering
\resizebox{\linewidth}{!}{
\begin{tikzpicture}[
  font=\footnotesize,
  box/.style={draw, rounded corners=2pt, align=center, minimum height=9mm, inner sep=3pt, fill=#1},
  box/.default=gray!8,
  arr/.style={-{Latex[length=2mm]}, thick},
  lab/.style={font=\scriptsize, align=center, text=black!70}]
  \node[box=blue!8] (img) {RGB frame\\(phone camera)};
  \node[box=blue!8, below=4mm of img] (vio) {6-DoF pose\\(VIO)};
  \node[box, right=6mm of img] (det) {person\\detection};
  \node[box, right=5mm of det] (lift) {ray--floor\\lifting};
  \node[box=orange!12, below=4mm of lift] (floor) {floor follows the\\smoothed camera\\height (\S\ref{sec:floor})};
  \node[box, right=5mm of lift] (trk) {world-frame\\Kalman tracker};
  \node[box, right=12mm of trk] (fc) {per-person heat-map\\forecast, 12 steps\\(Tier A / Tier B)};
  \node[box=green!10, right=6mm of fc] (rend) {re-projection\\into the\\current frame};
  \node[box, below=4mm of fc] (map) {bird's-eye map (0.2\,m cells):\\tracks, segmentation};
  \draw[arr] (img) -- (det);
  \draw[arr] (det) -- (lift);
  \draw[arr] (vio.east) -- (floor.west);
  \draw[arr] (floor) -- (lift);
  \draw[arr] (lift) -- (trk);
  \draw[arr] (trk) -- node[lab, above]{history} node[lab, below]{8 steps} (fc);
  \draw[arr] (map) -- (fc);
  \draw[arr] (trk.south) |- (map.west);
  \draw[arr] (fc) -- (rend);
  \draw[arr] (vio.south) -- ++(0,-4mm) -| (rend.south);
  \node[lab, above=1mm of fc] {up to 2.5\,Hz per person};
  \node[lab, above=1mm of det] {every frame};
\end{tikzpicture}}
\caption{System overview. Per frame, people are detected, their feet are
lifted onto the floor by intersecting pixel rays with a floor plane that
follows the phone's smoothed height (\cref{sec:floor}), and tracked in metres.
A bird's-eye map collects where tracked people walked and, at 2\,Hz, image
segmentation projected onto the floor (the RGB-to-map path is not drawn).
Each confirmed track is re-forecast at up to 2.5\,Hz as a probability
distribution over the ground for 12 future steps of 0.4\,s; the stored
forecasts are re-projected into every new frame with the current camera pose.}
\label{fig:system}
\end{figure}
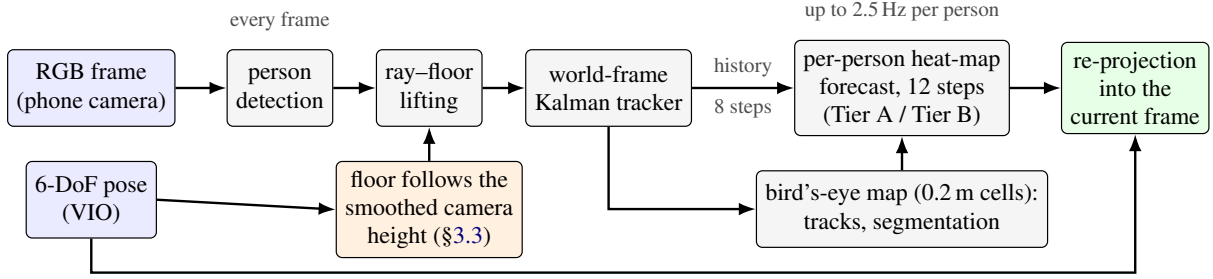

\Cref{fig:system} summarises the system. Everything that is forecast lives
in a gravity-aligned, metric world frame; the camera only decides where the
result is drawn.

\subsection{Setting and frames}
\label{sec:setting}
A phone records colour frames and, from its visual-inertial odometry (VIO;
ARKit on an iPhone), a 6-DoF camera pose per frame; the intrinsics are known.
We assume no depth sensor. Poses are expressed in a world frame $G$ with
$z$ pointing up (ARKit's world frame, which has $y$ up, is rotated
accordingly), in metres. Pedestrians are represented by their ground position
$\mathbf{x} = (x, y)$ in $G$.

\subsection{From detections to world-frame tracks}
\label{sec:lifting}
\paragraph{Detection.} A YOLOv8n person detector~\citep{jocher2023yolov8}
(confidence 0.35, non-maximum suppression at IoU 0.5, $640\times640$ input)
runs on the frame rotated upright according to the camera's roll, so that
phones held in portrait are handled. The foot point is the bottom centre of
the upright box.

\paragraph{Lifting.} The ray through the foot pixel is intersected with the
horizontal floor plane $z = z_f$ (\cref{sec:floor}). For a camera at height
$h$ above the floor and a ray depressed by $\theta$ below the horizon, the
ground distance is $d = h/\tan\theta$. A small error $\delta\theta$ in
attitude or foot pixel moves the point by approximately $\delta d \approx
(d^2/h)\,\delta\theta$, so we model the measurement noise as
\begin{equation}
\sigma(d) = \min\!\left(2\,\mathrm{m},\; 0.08\,\mathrm{m} + \frac{d^2}{h}\,\sigma_\theta\right),
\qquad \sigma_\theta = 0.0035\,\mathrm{rad},
\label{eq:noise}
\end{equation}
where $\sigma_\theta$ was set from the residuals of lifted positions against
smoothed tracks on a development clip (advio-12). Rays that meet the floor more
than 60\,m away are discarded. (With LiDAR depth, which our evaluation phone
lacks, the median of the torso's depth points is used instead, with
$\sigma = 0.10$\,m.)

\paragraph{Tracking.} Each person is tracked by a constant-velocity Kalman
filter~\citep{kalman1960new} in $G$ (white acceleration noise
$1\,\mathrm{m\,s^{-2}}$) with the measurement noise of \cref{eq:noise}.
Detections are assigned to tracks by the Hungarian algorithm on the
Mahalanobis distance, gated at the 99\,\% $\chi^2_2$ level; a track is
confirmed after three detections and deleted after 1\,s without one. Because
the state is the person's position on the ground, a standing person has a
constant state however the phone moves.

\subsection{Keeping the floor under a drifting phone}
\label{sec:floor}
Monocular lifting needs the floor's height in $G$. The obvious choice is to fix
the floor plane once, at $z_f = z_c(0) - h$ for the camera altitude $z_c$ at the
start. This is wrong whenever the reported altitude of the camera changes while
its height above the floor under the user does not: when the VIO drifts
vertically, and when the user walks up or down a slope or rides an escalator,
so that the floor under them changes height. With the floor fixed, a change
$\Delta z$ of the
reported camera altitude makes the camera appear $h + \Delta z$ above the floor,
and every lifted distance becomes
\begin{equation}
d' = \frac{h + \Delta z}{\tan\theta} = \frac{h + \Delta z}{h}\, d ,
\label{eq:scale}
\end{equation}
so positions and speeds are scaled by the same factor; when $\Delta z < -h$
no ray meets the floor and nobody can be placed. \Cref{sec:exp_drift} shows
that both happen on real recordings.

We instead keep the camera's \emph{height above the floor} fixed under its
low-pass-filtered altitude:
\begin{equation}
\bar z_c(t_k) = \bar z_c(t_{k-1}) + \left(1 - e^{-(t_k - t_{k-1})/\tau}\right)\bigl(z_c(t_k) - \bar z_c(t_{k-1})\bigr),
\qquad z_f(t_k) = \bar z_c(t_k) - \hat h ,
\label{eq:floor}
\end{equation}
with $\bar z_c(t_0) = z_c(t_0)$ and $\tau = 2$\,s, which removes the vertical
bob of walking but follows drift, slopes and escalators (with a lag of about
$v\tau$ while the altitude changes at a steady rate $v$; \cref{sec:discussion}).
The height $\hat h$ comes from LiDAR when available,
otherwise from a known value or from the people in view. For the latter we use
people as the ruler~\citep{hoiem2006putting,hoiem2008putting}: for an upright
person the ratio $r$ of head height to camera height follows from the camera's
attitude and the rays through the feet and the head alone, so with an assumed
adult height of 1.70\,m, $\hat h = 1.70\,\mathrm{m}/\operatorname{median}(r)$.
People whose feet are at least $8^\circ$ below the horizon (within about 10\,m)
are preferred, because distant people may stand on another level. In our
evaluation the height is estimated once per clip, offline, from the people
detected every 0.5\,s in the whole window, together with a correction of the
horizon (camera pitch) within $\pm3^\circ$: a pitch error biases each person's
estimate by an amount that grows with their distance, so the correction that
minimises the interquartile range of the log heights implied by all
detections is kept if it reduces that spread by more than
5\,\%~\citep{hoiem2008putting}. The correction is then applied to the camera
attitude of the whole clip, where it also absorbs errors of the intrinsic
calibration. Estimates outside 1.0--1.9\,m, with the optimum at the
$\pm3^\circ$ bound, or with a correction of more than $2.5^\circ$ are rejected
in favour of a 1.40\,m default without correction. (A live estimator of the
same ratio, without the pitch fit, runs when no height is given; it was not
used in this evaluation.) The rule of \cref{eq:floor} assumes that the people
being forecast stand on the same, locally level floor as the user; a slope
ahead of the user still biases far positions.

\subsection{Scene map}
\label{sec:map}
A bird's-eye class map with 0.2\,m cells (120\,m square, re-centred as the
user walks) accumulates weighted votes for six classes: unknown, pavement,
road, structure (impassable), terrain and tree. Cells where tracked people
walked receive walkable votes. With LiDAR, depth points vote for structure or
floor. Image segmentation (the public DeepLabv3+ model with an Xception-65
backbone~\citep{chen2017rethinking,chen2018encoder} trained on
ADE20K~\citep{zhou2017scene}, its classes mapped to our six) can add votes at
2\,Hz by projecting ground pixels through the floor plane (up to 35\,m, with
weights that fall with distance) and the ground contact of obstacles, which is
extended up to 6\,m behind a structure (with decaying weight) and 1\,m behind
vegetation, because ground hidden behind a wall is most likely more of the
same. Without depth this projection inherits every error of the floor
estimate, so whether it enters the forecasting map is a configuration choice
that we decided on development clips (\cref{sec:exp_phone}).

\subsection{Forecasters}
\label{sec:forecasters}
Every forecaster outputs, for each of the next 12 steps of 0.4\,s
($t = 0.4, \dots, 4.8$\,s), a probability distribution $p_t$ over the cells of a
$25.6\times25.6$\,m grid with 0.2\,m cells centred on the person, with zero
mass on impassable cells (before the hedge of \cref{eq:hedge}). Its input is
the person's last 8 filtered positions at 2.5\,Hz (2.8\,s; fewer if the track
is younger), the current and previous positions of the other tracked people,
and the local map.

\paragraph{Tier A: scene and physics prior (no training).}
Goals are drawn from a distribution over map cells that combines the walking
distance over the map (geodesic, so cells behind walls are far) relative to the
person's speed times the horizon, the agreement of the path's initial
direction with the current heading, a prior per map class, and a mild
penalty on detours. Samples then walk towards their goals in a social-force-like
rollout~\citep{helbing1995social} with path following, exponential repulsion
from walls and other people, acceleration noise, and hard limits on speed
($\le 2\,\mathrm{m\,s^{-1}}$ for walkers), acceleration
($\le 3\,\mathrm{m\,s^{-2}}$) and turning rate ($\le 3\,\mathrm{rad\,s^{-1}}$);
samples that would enter an obstacle are projected back. A kernel density
estimate of the samples (96 in the live pipeline and on EgoTraj-Bench, 128 on
SDD), with a bandwidth that grows with the horizon, gives $p_t$. The class
prior is the observed enrichment of future positions on each class in the SDD
training windows; the nine other parameters and the acceleration noise were
chosen by a random search over bounded ranges (97 configurations, then 48
perturbations of the best) that maximises the likelihood of the true positions
at 1.2, 2.4 and 4.8\,s on training windows only: 800 from SDD (clean tracks)
and, for a second profile that keeps the SDD class prior and detour weight,
1,200 from EgoTraj-Bench (noisy first-person tracks).

\paragraph{Tier B: learned heat map.}
A U-Net~\citep{ronneberger2015unet} with five encoder levels (widths 32, 48,
64, 96 and 128; 1.02\,M parameters) maps a $128\times128$ crop at 0.2\,m to
12 per-step maps, each normalised by a spatial softmax. The crop is centred on
the person's current position, rotated so that their heading points up, with
the person 30\,\% of the crop height from the back edge so that most of the
crop lies ahead. The 16 input channels are the six map classes (one-hot,
including \emph{unknown}), one Gaussian blob per past position, and the
current and previous positions of the neighbours. The network is trained with
the negative log-likelihood of the true future cell, softened by a Gaussian of
1.5 cells, so its outputs are distributions rather than per-pixel
probabilities as in Y-Net~\citep{mangalam2021goals}. Training data are the
pedestrian trajectories of the SDD TrajNet training split with Y-Net's
hand-labelled semantic maps, converted to metres with OpenTraj's per-video
scales~\citep{amirian2021opentraj}, and all \TrainWindowsEgo{} training windows of
EgoTraj-Bench (TBD, v1.1)~\citep{liu2026egotraj}, whose real first-person
histories are noisy and which have no map (all cells \emph{unknown}). To
resemble what a phone sees, training applies field-of-view visibility masks
(a simulated camera with a $60$--$75^\circ$ field of view 4--14\,m from the
person; outside is \emph{unknown}) with probability 0.6, history jitter,
truncated histories, occasional identity-switch jumps, heading noise and
mirroring (\cref{app:training}). At inference the crop is filled from the
person's $25.6\times25.6$\,m local map (beyond it: \emph{unknown}), the output
is resampled onto that grid, masked by the impassable cells and renormalised.
A person who has moved less than 0.15\,m over the last three steps is given a
fixed heading, and in the live pipeline a neighbour's previous position is
extrapolated back by 0.4\,s with its tracked velocity. The predicted
last-step map can also serve as the goal distribution of the Tier A rollout
(``Tier B goals'').

\paragraph{Combination, hedge and position uncertainty.}
With the network loaded, the forecast mixes its heat map $p^{\mathrm{net}}_t$
with the physics rollout steered by its goals, $p^{\mathrm{goals}}_t$:
$p_t = (1-w)\,p^{\mathrm{goals}}_t + w\,p^{\mathrm{net}}_t$, with $w$ chosen on
validation data (\cref{sec:exp_fusion}); the choice was $w = 1$, so the
rollout is skipped. The
forecast is then convolved with an isotropic Gaussian of the tracker's current
position uncertainty (re-masked and renormalised) and mixed with a small weight
$\varepsilon$ of a constant-velocity Gaussian whose per-step width was fitted on
training data:
\begin{equation}
\tilde p_t = (1-\varepsilon)\,\bigl(p_t * \mathcal N(0, \sigma_{\mathrm{pos}}^2 I)\bigr)
          + \varepsilon\, \mathcal N\!\bigl(\mathbf{x}_0 + t\,\mathbf{v}_0,\; s_t^2 I\bigr),
\label{eq:hedge}
\end{equation}
where $\mathbf{x}_0, \mathbf{v}_0$ are the last position and a least-squares
velocity over the last 1.2\,s. This hedge, not masked by the map, bounds the
damage of map and model errors. Two parameter profiles exist: one fitted on
clean SDD tracks ($\varepsilon = 0.15$) for people whose recent measurement
noise is at most 0.12\,m, and one fitted on EgoTraj-Bench's noisy first-person
tracks ($\varepsilon = 0.25$) for everyone else. Without LiDAR, by
\cref{eq:noise}, only people within about 4\,m qualify for the first.

\subsection{Rendering into the moving view}
\label{sec:render}
Forecasts are stored on their world-aligned grids. For display, each step's
50\,\% highest-density region is drawn on the floor and warped into the
current frame with the floor homography of the current camera, so the
overlay stays attached to the ground while the phone moves; non-ground pixels
(from segmentation) are not painted. Display choices do not affect any score
in this paper.

\section{Evaluation methodology}
\label{sec:evaluation}

\subsection{Benchmarks and metrics}
\label{sec:benchmarks}
\paragraph{Stanford Drone Dataset (SDD).} We use the standard TrajNet split of
SDD~\citep{robicquet2016learning,sadeghian2018trajnet} as distributed with
Y-Net~\citep{mangalam2021goals}: 8 observed and 12 future positions at 2.5\,Hz,
with Y-Net's hand-labelled semantic maps. Positions are converted to metres
with OpenTraj's per-video scale estimates~\citep{amirian2021opentraj}, some of
which are themselves estimates. All \SddTestN{} test trajectories are
evaluated. Five whole scenes of the training split (\SddValScenes{};
\SddValN{} windows) are held out as validation data for the network and are
never trained on; the Tier A parameters, the hedge weight and the CV widths,
all fitted earlier on the whole training split, did see them. Seen from a
drone, roofs and arcades hide walkable ground: \SddTrainOnStructure{} of the
true future positions in the training split lie on cells labelled as
structure, most of them near its edge. On SDD, all methods therefore treat
only structure more than 2\,m from its edge as impassable (a choice made while
developing Tier A; \SddTrainOnCore{} of the training positions lie there); on
the phone, all structure is impassable.

\paragraph{EgoTraj-Bench.} The TBD subset of EgoTraj-Bench (data release
v1.1)~\citep{liu2026egotraj,wang2024tbd} pairs noisy first-person histories
with clean bird's-eye futures, with the same 8/12 steps at 2.5\,Hz and a
chronological split (\EgoTrainN{} training, \EgoValN{} validation and
\EgoTestN{} test windows in our loader, which interpolates invalid history
steps and skips windows with fewer than two valid ones). There is no map.

\paragraph{Metrics.} The primary score is the negative log-likelihood (NLL)
of the true position under the forecast distribution at 1.2, 2.4 and 4.8\,s,
as a density per m$^2$ on the 0.2\,m grid, $-\log\bigl(\max(p_t(\text{cell}),
10^{-6})/0.04\,\mathrm{m^2}\bigr)$, in nats; the floor caps the penalty of a
miss at 10.6~nats. Without the floor it is a strictly proper
score~\citep{gneiting2007strictly}.
Calibration is measured by the coverage of the $\alpha$-highest-density
region~\citep{hyndman1996computing} (the share of truths inside it, ideally
$\alpha$) and sharpness by the area of the 90\,\% region. For comparison with
published work we also give best-of-20 displacement errors (minADE$_{20}$,
minFDE$_{20}$) of 20 random samples. These exist only for sample-based
forecasters (Tier A, and Tier B goals with the physics rollout); the final
heat-map forecaster has no trajectories. On SDD they are also given in pixels
of the original images, the unit of the published results.

\paragraph{Baseline.} The constant-velocity (CV) Gaussian extrapolates the
least-squares velocity of the last 1.2\,s and places an isotropic Gaussian with
the per-step width fitted by maximum likelihood on the training split of the
same benchmark. It is the reference in every comparison: a strong
baseline~\citep{scholler2020what} whose spread is matched to the training
data, although a Gaussian cannot be calibrated at every level when the errors
are heavy-tailed (\cref{sec:exp_explore} examines this on phone video).

\subsection{Real handheld-phone video without ground truth}
\label{sec:selfconsistency}
\paragraph{Recordings.} No public recording of pedestrians from a handheld
phone with ground-truth pedestrian trajectories exists that we are aware of.
ADVIO~\citep{cortes2018advio} provides 23 sequences recorded by a person
walking with a hand-held test rig that carries an iPhone~6s (and two other
devices), with the iPhone's ARKit poses, reference (``ground-truth'') poses
computed offline from inertial navigation with manual position fixes, and
crowds in a mall, a metro station and city streets, but no labels for the
people. We use 12 sequences with people (\cref{tab:clips}). For each, a
converter picks the 90\,s window with the most detected people, rotates the
frames upright, removes lens distortion and writes frames at 30\,fps (ADVIO
records at 60) with poses in ARKit's convention. ADVIO provides three
per-session calibrations of the camera but does not say which session
recorded which sequence; we use the second for all clips (their principal
points differ by up to $1.3^\circ$, an error that the horizon correction of
\cref{sec:floor} can absorb where it is applied, which is not on the two
clips whose people-based estimate is rejected). The converter uses ARKit's
poses unless they
fail a plausibility check over the whole sequence (speed, jumps, pitch), in
which case ADVIO's reference poses are used: on one development clip (23) and
on three held-out clips (02, 04, 05), which carry \PoseRefShareHeld{} of the
held-out forecasts. These clips are therefore not phone-only; results by pose
source are in \cref{sec:exp_explore}. The camera height and the horizon
correction are estimated from the people in the window (\cref{sec:floor}).

\paragraph{Self-consistency scoring.} Every forecast, made at time $t_0$ for a
track, is scored at $t_0 + 1.2$, $2.4$ and $4.8$\,s against where the tracker
later measured that person, in two ways:
\begin{enumerate}[leftmargin=*]
\item \emph{Raw truth} (primary): the single measurement nearest the target
time (within 0.1\,s), with both the forecast and the baseline convolved with
that measurement's noise $\sigma(d)$ of \cref{eq:noise}, i.e., the likelihood of
what was actually measured.
\item \emph{Smoothed truth}: the track's Rauch--Tung--Striebel
smoother~\citep{rauch1965maximum} with 2\,s of look-ahead, evaluated at the
target time.
\end{enumerate}
The smoother assumes constant velocity, so where measurements are noisy its
``truth'' is pulled towards straight lines, which can favour the CV baseline.
The raw truth is not smoothed, which is why we make it primary (on the
development clips it also turned out to be the more conservative of the two,
but not on the held-out clips; \cref{sec:exp_phone,sec:exp_heldout}); it is
not free of the motion model either, because measurements are associated to a
track through the gate of a constant-velocity Kalman filter, so the targets
that survive are consistent with constant velocity to within the gate. A
forecast is scored against the raw truth only if a measurement lies within
0.1\,s of the target time, so fewer forecasts are scored against it than
against the smoothed truth. The CV baseline is scored on the same forecasts;
its per-step widths are those fitted on the training split behind the person's
profile (SDD or EgoTraj-Bench; there are no phone training data), widened by
the same current position uncertainty as our forecast. Forecasts are grouped
by the person's distance from the camera when the forecast was made ($<8$,
8--15 and $>15$\,m), because the lifting noise grows with the square of the
distance (\cref{eq:noise}).

\paragraph{What this measures, and what it does not.} The pseudo truth is
the system's own later measurement. It inherits lifting errors (which are
shared by the forecaster's input and its truth), identity switches, and
survivorship: only people who are still tracked at the horizon are scored. A
forecaster can therefore look good by predicting where the \emph{tracker}
will put a person rather than where the person will be. Both compared
forecasters see the same tracks and are scored on the same targets, so the
comparison between them is paired; absolute numbers are not comparable with
the benchmarks.

\subsection{Statistics}
\label{sec:stats}
All comparisons are paired: two methods are scored on the same windows or
the same forecasts (tracking does not depend on the forecaster, so variants
of a clip that differ only in the forecaster are scored on identical
forecasts; we verified this). We
report mean differences with 95\,\% percentile bootstrap intervals from
2,000 resamples~\citep{davison1997bootstrap}, rounded to two decimals (an
interval bound printed as $-0.00$ or $+0.00$ lies just below or above zero).
Rows that belong together are
resampled together: on phone video, all forecasts of one person (clip and
track identity; a person whose track breaks counts as several, which makes
these intervals somewhat too narrow); on SDD, each window (each is a different
trajectory) and, as
a check, each scene; on EgoTraj-Bench, each window and, as a check, each
group of simultaneously observed agents. Intervals over people treat the
phone clips themselves as fixed; \cref{sec:exp_explore} also resamples clips.

\subsection{Protocol}
\label{sec:protocol}
Five ADVIO clips (06, 11, 12, 22, 23) were used for development: sensor-model,
tracker, map, display and configuration decisions, and the floor fix of
\cref{sec:floor} were all made while looking at them. Seven further clips
(01--05, 20, 21) were downloaded and converted only after every other
decision had been made; the last one, whether segmentation enters the
forecasting map, was then taken on the development clips by a rule written
into the protocol. Before any forecast was run on the seven clips, we
committed this protocol (\cref{app:protocol}), which fixes the configuration,
the weights (by checksum), the primary comparison and the analysis, in the
spirit of pre-registration~\citep{nosek2018preregistration}. The
pre-registration is internal: a commit in our own repository, with its hash
and time but no external timestamp. The analyses it specifies are reported in
full, and comparisons it did not specify are marked as such. The benchmark test
splits were evaluated several times during development (SDD \SddTestUses{}
times and EgoTraj-Bench \EgoTestUses{} times in total, including the final
runs reported here); every choice between those runs was made on training or
validation data or was a bug fix, but some selection effect cannot be ruled
out.

\section{Experiments}
\label{sec:experiments}

\subsection{Benchmarks}
\label{sec:exp_bench}

\begin{table}[t]
\centering\small\setlength{\tabcolsep}{4.5pt}
\begin{tabular}{@{}lrrrrrrrr@{}}
\toprule
 & \multicolumn{3}{c}{NLL (nats) $\downarrow$} & \multicolumn{2}{c}{coverage @4.8\,s} & area$_{90}$ & \multicolumn{2}{c}{best of 20 (m) $\downarrow$} \\
\cmidrule(lr){2-4}\cmidrule(lr){5-6}\cmidrule(lr){8-9}
Method & 1.2\,s & 2.4\,s & 4.8\,s & 50\,\% & 90\,\% & (m$^2$) & ADE & FDE \\
\midrule
\multicolumn{9}{@{}l}{\emph{Stanford Drone, TrajNet test (2,829 windows)}} \\
CV Gaussian & 0.72 & 2.03 & 3.68 & 0.85 & 0.95 & 56.6 & 0.59$^\ast$ & 1.19$^\ast$ \\
Tier A (physics prior) & 0.35 & 1.57 & 3.09 & 0.60 & 0.90 & 26.7 & 0.49 & 0.75 \\
Tier B goals + physics & \textbf{0.25} & 1.35 & 2.54 & 0.59 & 0.89 & 14.7 & 0.47 & 0.68 \\
Tier B heat map, no hedge & 0.37 & 1.07 & 2.20 & 0.59 & 0.86 & 12.3 & -- & -- \\
Tier B heat map (final) & 0.28 & \textbf{1.02} & \textbf{2.17} & 0.65 & 0.93 & 22.8 & -- & -- \\
\quad heat map $-$ CV & \makecell[r]{$-0.44$\\[-1pt]{\scriptsize[$-0.46$,\,$-0.42$]}} & \makecell[r]{$-1.01$\\[-1pt]{\scriptsize[$-1.05$,\,$-0.98$]}} & \makecell[r]{$-1.51$\\[-1pt]{\scriptsize[$-1.57$,\,$-1.46$]}} & \multicolumn{5}{l}{\scriptsize (95\,\% interval over windows)} \\
\midrule
\multicolumn{9}{@{}l}{\emph{EgoTraj-Bench TBD, test (9,828 windows)}} \\
CV Gaussian & 1.77 & 2.82 & 4.31 & 0.67 & 0.83 & 58.8 & 1.17$^\ast$ & 2.21$^\ast$ \\
Tier A (physics prior) & 1.63 & 2.72 & 4.13 & 0.66 & 0.84 & 50.3 & 0.78 & 1.29 \\
Tier B goals + physics & 1.38 & 2.26 & 3.53 & 0.72 & 0.92 & 43.6 & 0.59 & 0.88 \\
Tier B heat map, no hedge & \textbf{0.66} & \textbf{1.50} & \textbf{2.86} & 0.48 & 0.84 & 16.8 & -- & -- \\
Tier B heat map (final) & 0.78 & 1.60 & 2.94 & 0.58 & 0.93 & 37.0 & -- & -- \\
\quad heat map $-$ CV & \makecell[r]{$-0.99$\\[-1pt]{\scriptsize[$-1.01$,\,$-0.96$]}} & \makecell[r]{$-1.22$\\[-1pt]{\scriptsize[$-1.25$,\,$-1.20$]}} & \makecell[r]{$-1.37$\\[-1pt]{\scriptsize[$-1.40$,\,$-1.34$]}} & \multicolumn{5}{l}{\scriptsize (95\,\% interval over windows)} \\
\bottomrule
\end{tabular}

\caption{Test-set results. NLL of the true position (nats; lower is better),
coverage of the 50\,\% and 90\,\% highest-density regions at 4.8\,s (calibrated
forecasts reach the nominal level), area of the 90\,\% region at 4.8\,s, and
best-of-20 errors over the 4.8\,s horizon (only for sample-based
forecasters; $^\ast$single path). Tier A, Tier B goals and the final
configuration, the Tier B heat map, include the constant-velocity hedge of
\cref{eq:hedge} ($\varepsilon=0.15$ on SDD, $0.25$ on EgoTraj-Bench). Bold:
best NLL. The last row of each block is the paired difference between the
final configuration and the CV baseline with a 95\,\% bootstrap interval.}
\label{tab:bench}
\end{table}

\Cref{tab:bench} compares the forecasters on the two test splits. The final
forecaster, the network's heat map with the hedge of \cref{eq:hedge}, improves
on the CV baseline at every horizon, on SDD by \BenchSddGainOne{},
\BenchSddGainTwo{} and \BenchSddGainFour{}~nats at 1.2, 2.4 and 4.8\,s and on
EgoTraj-Bench by \BenchEgoGainOne{}, \BenchEgoGainTwo{} and
\BenchEgoGainFour{}~nats, with bootstrap intervals far from zero also when
whole SDD scenes or EgoTraj-Bench groups are resampled (at 4.8\,s:
\BenchSddGainFourClusterCI{} and \BenchEgoGainFourClusterCI{}), and on Tier A by
\BenchSddGainAFour{} and \BenchEgoGainAFour{}~nats at 4.8\,s. Using the network
only to choose goals for the physics rollout (``Tier B goals'') captures part
of the gain; the heat map itself is better at every horizon on EgoTraj-Bench
and at 2.4 and 4.8\,s on SDD, by \BenchSddGainGoalsFour{} (SDD) and
\BenchEgoGainGoalsFour{} (EgoTraj-Bench) nats at 4.8\,s, while at 1.2\,s on
SDD the goal-conditioned rollout has a slightly lower NLL (heat map $-$ goals:
\BenchSddGoalsOneWindowCI{} over windows, \BenchSddGoalsOneClusterCI{} over
scenes). The constant-velocity hedge of \cref{eq:hedge} trades
sharpness for coverage: without it the network's heat map has smaller 90\,\%
regions (\BenchSddNetAloneArea{} instead of \BenchSddHeatArea{}\,m$^2$ on SDD
at 4.8\,s) and, on EgoTraj-Bench, a lower NLL (\BenchEgoNetAloneNllFour{}
instead of \BenchEgoHeatNllFour{} at 4.8\,s), but there it is over-confident
(90\,\% coverage \BenchEgoNetAloneCovNinetyFour{}); on SDD the hedge improves
the NLL at every horizon. The hedge weights were fitted for Tier A and not
re-tuned for the network.

On best-of-20 displacement, the sample-based variants are far from the state
of the art (\cref{tab:published}): Tier B goals reaches
\BenchSddGoalsADEpx{}/\BenchSddGoalsFDEpx{} pixels on SDD, better than
Social GAN and SoPhie but far behind PECNet and Y-Net (7.85/11.85
pixels~\citep{mangalam2021goals}), and \BenchEgoGoalsADE{}/\BenchEgoGoalsFDE{}\,m
on EgoTraj-Bench, against about 0.30/0.44\,m for BiFlow's released
checkpoints~\citep{liu2026egotraj}. Our forecasters were built and selected
for calibrated densities, not for this metric, which rewards spread-out
samples~\citep{thiede2019analyzing}.

\subsection{Choosing the fusion weight, and training variance}
\label{sec:exp_fusion}
The weight $w$ of the network's own heat map against the physics rollout
steered by the network's goals was chosen per profile on validation data only:
the SDD validation scenes for the clean-track profile and EgoTraj-Bench
validation for the noisy one. The mean NLL over the three horizons falls
monotonically as $w$ grows, from \FusionValSddWZero{} ($w=0$) to
\FusionValSddWOne{} ($w=1$) on SDD and from \FusionValEgoWZero{} to
\FusionValEgoWOne{} on EgoTraj-Bench (\cref{fig:fusion}), so $w=1$ was chosen
for both: the final forecaster is the network's heat map with the hedge of
\cref{eq:hedge}, and the rollout is skipped. Two biases act on this choice in
opposite directions: Tier A's parameters were fitted on all SDD training
scenes, the validation scenes included, which favours the physics, while the
network was early-stopped on the same validation sets, which favours the
network. Per horizon the choice is not uniformly best: at 1.2\,s on SDD a
mixture ($w=0.4$) has the lowest validation NLL.

\begin{table}[t]
\centering\small
\begin{tabular}{@{}lrrrrrrrrr@{}}
\toprule
 & & & & \multicolumn{3}{c}{SDD test NLL} & \multicolumn{3}{c}{EgoTraj test NLL} \\
\cmidrule(lr){5-7}\cmidrule(lr){8-10}
Seed & $w$ (SDD / Ego) & best epoch & val.\ score & 1.2\,s & 2.4\,s & 4.8\,s & 1.2\,s & 2.4\,s & 4.8\,s \\
\midrule
0 & 1.0 / 1.0 & 46 & 1.338 & 0.28 & 1.02 & 2.17 & 0.78 & 1.60 & 2.94 \\
1 & 1.0 / 1.0 & 30 & 1.358 & 0.27 & 1.04 & 2.20 & 0.82 & 1.65 & 2.96 \\
2 & 1.0 / 1.0 & 34 & 1.343 & 0.24 & 1.02 & 2.17 & 0.80 & 1.63 & 2.95 \\
\midrule
mean & & & 1.346 & 0.26 & 1.03 & 2.18 & 0.80 & 1.63 & 2.95 \\
s.d. & & & 0.010 & 0.02 & 0.01 & 0.02 & 0.02 & 0.03 & 0.01 \\
CV Gaussian & & & & 0.72 & 2.03 & 3.68 & 1.77 & 2.82 & 4.31 \\
\bottomrule
\end{tabular}

\caption{Training variance: the final configuration trained with three seeds
(seed 0 is the network used everywhere else). For each seed, the fusion weight
$w$ was chosen on its own validation runs and the test sets were scored once
with it. ``val.\ score'': the checkpoint-selection score at the best epoch
(mean of the SDD and EgoTraj-Bench validation NLLs, \cref{app:training}).
Test NLL of the final configuration (with the hedge), in nats.}
\label{tab:seeds}
\end{table}

On the benchmarks, training is stable across seeds (\cref{tab:seeds}; the
phone results use seed 0 only). The validation runs of the two further seeds
chose the same fusion weight as seed 0, $w = 1$ for both profiles, and the
final configuration's test NLL differs between the three seeds by at most
\SeedsSddRangeMax{} (SDD) and \SeedsEgoRangeMax{} (EgoTraj-Bench) nats at any
horizon, far less than its distance to the CV baseline.

\subsection{Vertical drift on real recordings}
\label{sec:exp_drift}

\begin{figure}[t]
\centering
\includegraphics[width=\linewidth]{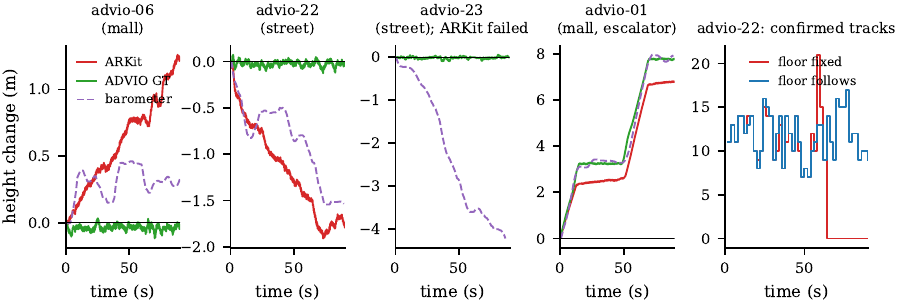}
\caption{Left four panels: change of the phone's height over the 90\,s clip
according to ARKit, ADVIO's reference poses (``GT'') and the phone's
barometer. advio-06: ARKit drifts upward by \DriftSixArkitAbs{} while the
reference poses stay level (barometer \DriftSixBaro{}, within its noise).
advio-22: ARKit and the barometer agree on a descent of about 1.5--1.8\,m along
a sloping street that the reference poses do not show. advio-23: ARKit fails;
the barometer indicates a descent of about 4\,m that the reference poses do not
show. advio-01 (held out): two escalator rides. Right: confirmed tracks on
advio-22 in the development run of Tier A with the floor held at its
first-frame height and with the floor following the phone (\cref{sec:floor}).}
\label{fig:drift}
\end{figure}

\Cref{fig:drift} and \cref{tab:clips} show how the phone's altitude in the
VIO frame changes over the clips. On advio-06, a mall with a level floor,
ARKit's height rises by \DriftSixArkitAbs{} in 90\,s while ADVIO's reference
poses stay level (\DriftSixGT{}). With the floor plane held at its first-frame
height, the camera therefore appears \InflationSixEnd{} higher above the floor
at the end of the clip than at the start, and by \cref{eq:scale} every lifted
distance and speed is inflated by the same factor. On advio-22, a street that
descends by about 1.5\,m according to the barometer (\SlopeTwentyTwoBaro{};
ARKit \SlopeTwentyTwoArkit{}, the reference poses stay level), the reported
height eventually drops below the fixed floor: no ray meets it, and the
tracker loses everyone for the last \CollapseShare{} of the clip although
people remain in view (\cref{fig:drift}, right). Escalators change the true
height by metres (advio-01: \EscalatorOneGT{}; advio-04: \EscalatorFourGT{}).
Note also that ADVIO's reference poses stay level on both street clips where
the barometer indicates a descent (\SlopeTwentyThreeBaro{} on advio-23), so
they cannot serve as a vertical reference outdoors.

\Cref{tab:floor_dev} repeats the development runs of Tier A with the floor held
fixed. Where the reported altitude changes, the fix changes the geometry
substantially: on advio-06, where the reported altitude rises, the share of
forecasts made for people within 8\,m rises from \FloorSixNearFixed{} to
\FloorSixNearFollow{} (distances are no longer inflated); on advio-11, where it
falls by \DriftElevenArkitAbs{}, the fixed floor compresses distances and puts
\FloorElevenNearFixed{} of the forecasts within 8\,m instead of
\FloorElevenNearFollow{}; on advio-22 it puts \FloorTwentyTwoNearFixed{} there
instead of \FloorTwentyTwoNearFollow{} before it loses everyone, and the time
without any track falls from \CollapseShare{} to \FloorTwentyTwoNoTrackFollow{},
with \CollapseNFollow{} instead of \CollapseNFixed{} forecasts scored; on
advio-23, whose reference poses are level, the counts and the near share do not
change. The self-consistency score cannot show whether the rescaled positions
are wrong (a wrong floor rescales the input and the truth alike); the
synthetic test of \cref{app:synthetic} isolates the effect with known ground
truth.

\begin{table}[t]
\centering\small
\begin{tabular}{@{}lrrrrrrrr@{}}
\toprule
 & \multicolumn{2}{c}{forecasts scored} & \multicolumn{2}{c}{within 8\,m} & \multicolumn{2}{c}{time with no track} & \multicolumn{2}{c}{NLL $-$ CV @1.2\,s} \\
\cmidrule(lr){2-3}\cmidrule(lr){4-5}\cmidrule(lr){6-7}\cmidrule(lr){8-9}
Clip & fixed & follows & fixed & follows & fixed & follows & fixed & follows \\
\midrule
advio-06 & 717 & 770 & 7\,\% & 20\,\% & 9\,\% & 7\,\% & $-0.18$ & $-0.08$ \\
advio-11 & 583 & 568 & 32\,\% & 20\,\% & 20\,\% & 20\,\% & $-0.09$ & $-0.20$ \\
advio-12 & 533 & 543 & 1\,\% & 1\,\% & 11\,\% & 11\,\% & $-0.16$ & $-0.12$ \\
advio-22 & 803 & 1,107 & 57\,\% & 7\,\% & 31\,\% & 0\,\% & $+0.42$ & $-0.04$ \\
advio-23 & 147 & 147 & 7\,\% & 7\,\% & 20\,\% & 20\,\% & $+0.04$ & $0.00$ \\
\bottomrule
\end{tabular}

\caption{Floor held fixed at the first frame (as before the fix) versus following
the phone, on the development clips (Tier A). Forecasts scored at 1.2\,s,
the share of them made for people within 8\,m, the share of the clip
(sampled every 2\,s) with no confirmed track, and the mean NLL difference to
the CV baseline at 1.2\,s (raw truth). The NLL differences of the two variants
are not directly comparable, because a wrong floor rescales positions and
truth alike.}
\label{tab:floor_dev}
\end{table}

\subsection{Real handheld video: development clips}
\label{sec:exp_phone}
\Cref{tab:phone_dev} pools the \DevNOne{} forecasts scored at 1.2\,s on the
five development clips (\DevPeopleOne{} tracked people). The final forecaster,
the network's heat map with segmentation in the map (the map decision, made
here by the rule of \cref{app:protocol}, improved the 1.2\,s NLL by
\DevMapEffectOne{} \DevMapEffectOneCI{} nats), has a lower NLL than the CV
baseline by \DevGainOne{} \DevGainOneCI{}, \DevGainTwo{} \DevGainTwoCI{} and
\DevGainFour{} \DevGainFourCI{} nats at 1.2, 2.4 and 4.8\,s, and lower than
Tier A by \DevGainAOne{}, \DevGainATwo{} and \DevGainAFour{} nats. It is better
in every distance band and, by point estimate, on every clip (significantly at
all three horizons on \DevClipsSigAll{} of the five;
\cref{tab:phone_dev,tab:perclip_dev}). Its 90\,\% regions
contain the raw measurement \DevCovNinetyOne{}, \DevCovNinetyTwo{} and
\DevCovNinetyFour{} of the time (CV: \DevCovNinetyCVOne{}, \DevCovNinetyCVTwo{},
\DevCovNinetyCVFour{}). Against the smoothed truth the advantage is larger
on these clips (\DevGainSmoothOne{} nats at 1.2\,s), so here the raw score is
the conservative one; on the held-out clips it is not
(\cref{sec:exp_heldout}). Tier A, the
physics prior, is only marginally better than CV (by \DevGainTierACVOne{} nats
at 1.2\,s), and not at all within 8\,m. These clips informed many design
decisions, so the numbers are optimistic; \cref{sec:exp_heldout} tests the
frozen system on new clips.

\begin{table}[t]
\centering\small
\begin{tabular}{@{}llrrrrll@{}}
\toprule
Ahead & Distance & $n$ (people) & final & Tier A & CV & final $-$ CV & final $-$ Tier A \\
\midrule
1.2\,s & all & 3,135 (462) & 2.59 & 2.93 & 3.02 & $-0.43$ {\scriptsize[$-0.49$, $-0.38$]} & $-0.34$ {\scriptsize[$-0.40$, $-0.30$]} \\
 & <8 m & 361 (106) & 0.95 & 1.54 & 1.54 & $-0.59$ {\scriptsize[$-0.72$, $-0.49$]} & $-0.59$ {\scriptsize[$-0.76$, $-0.44$]} \\
 & 8--15 m & 1,069 (222) & 1.67 & 1.98 & 2.04 & $-0.37$ {\scriptsize[$-0.44$, $-0.30$]} & $-0.32$ {\scriptsize[$-0.39$, $-0.25$]} \\
 & >15 m & 1,705 (298) & 3.51 & 3.82 & 3.95 & $-0.44$ {\scriptsize[$-0.52$, $-0.37$]} & $-0.31$ {\scriptsize[$-0.39$, $-0.24$]} \\
\addlinespace
2.4\,s & all & 2,137 (311) & 3.45 & 4.17 & 4.32 & $-0.86$ {\scriptsize[$-0.96$, $-0.77$]} & $-0.72$ {\scriptsize[$-0.81$, $-0.63$]} \\
 & <8 m & 158 (42) & 1.74 & 2.65 & 2.61 & $-0.86$ {\scriptsize[$-1.10$, $-0.64$]} & $-0.91$ {\scriptsize[$-1.15$, $-0.66$]} \\
 & 8--15 m & 703 (143) & 2.40 & 2.96 & 3.02 & $-0.62$ {\scriptsize[$-0.78$, $-0.48$]} & $-0.57$ {\scriptsize[$-0.71$, $-0.44$]} \\
 & >15 m & 1,276 (221) & 4.25 & 5.03 & 5.24 & $-0.99$ {\scriptsize[$-1.14$, $-0.86$]} & $-0.78$ {\scriptsize[$-0.92$, $-0.66$]} \\
\addlinespace
4.8\,s & all & 997 (152) & 4.70 & 5.73 & 5.90 & $-1.21$ {\scriptsize[$-1.39$, $-1.03$]} & $-1.03$ {\scriptsize[$-1.19$, $-0.87$]} \\
 & <8 m & 48 (9) & 2.58 & 3.55 & 3.60 & $-1.02$ {\scriptsize[$-1.57$, $-0.41$]} & $-0.97$ {\scriptsize[$-1.32$, $-0.49$]} \\
 & 8--15 m & 276 (48) & 3.30 & 3.92 & 4.20 & $-0.90$ {\scriptsize[$-1.25$, $-0.53$]} & $-0.62$ {\scriptsize[$-0.88$, $-0.32$]} \\
 & >15 m & 673 (119) & 5.42 & 6.63 & 6.77 & $-1.34$ {\scriptsize[$-1.55$, $-1.14$]} & $-1.21$ {\scriptsize[$-1.41$, $-1.02$]} \\
\bottomrule
\end{tabular}

\caption{Development clips (advio-06, 11, 12, 22, 23), pooled. NLL against
the raw measurement nearest each horizon, by the person's distance when the
forecast was made; $n$ forecasts from the given number of people. Differences
are paired, with 95\,\% intervals from resampling people.}
\label{tab:phone_dev}
\end{table}

\subsection{Real handheld video: pre-registered held-out clips}
\label{sec:exp_heldout}

\begin{figure}[t]
\centering
\includegraphics[width=\linewidth]{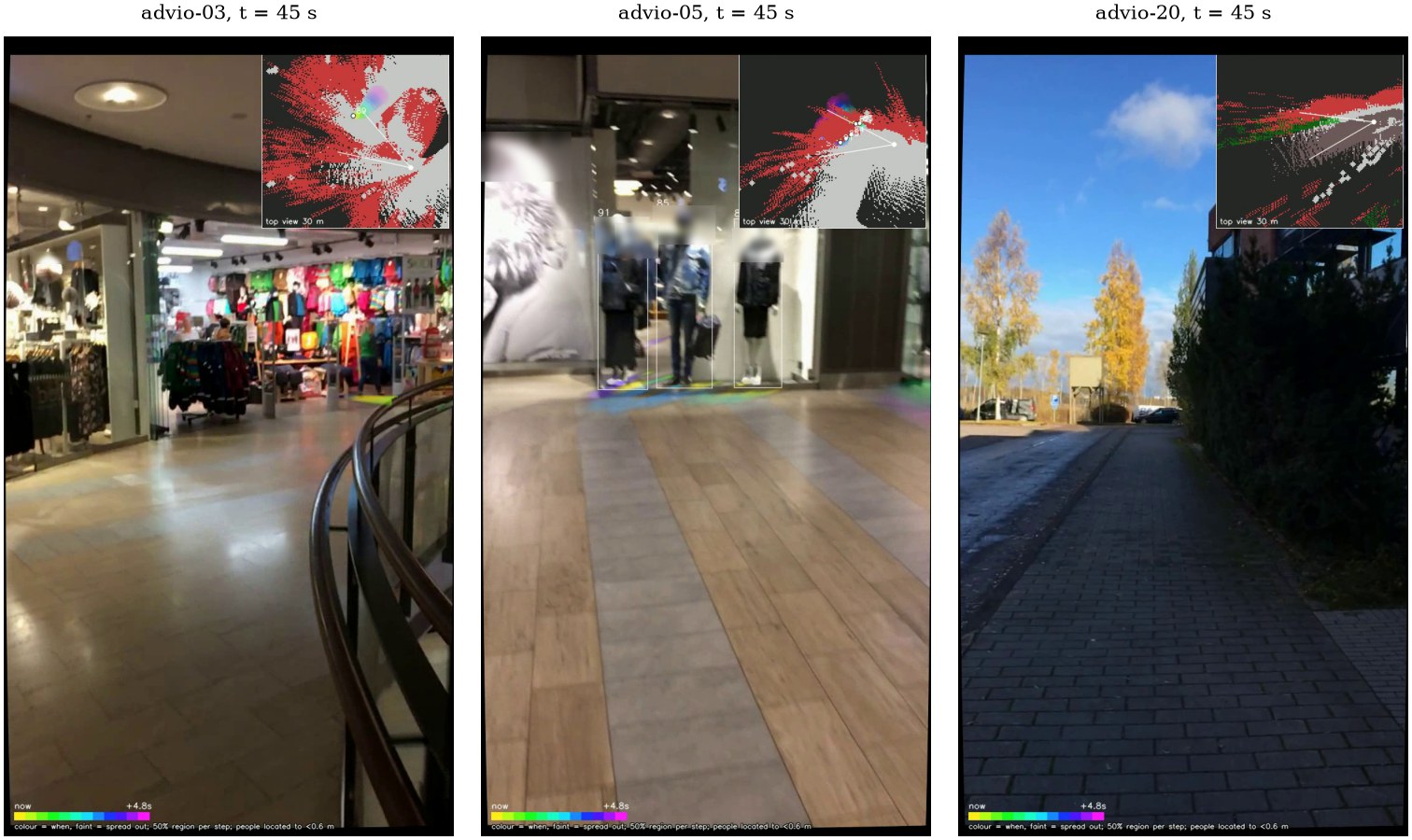}
\caption{The final forecaster on three held-out clips at the middle of each
clip ($t=45$\,s); the clips and the time were fixed in the protocol before the
runs, so the frames are not selected. Coloured regions on the floor are the
50\,\% highest-density regions of each person's forecast, from 0.4\,s ahead
(yellow) through green and blue to 4.8\,s ahead (magenta), drawn fainter where
the forecast is spread out; the inset is the bird's-eye view (30\,m). At this
instant advio-03 and advio-20 have nobody close by, and in advio-05 shop-window
mannequins are detected as people and forecast. Forecasts of people whose
position is uncertain by more than 0.6\,m are not drawn (display only;
\HiddenOverlayHeld{} of the scored forecasts). Selected frames with more
people are in \cref{fig:frames_selected}. Frames from ADVIO (CC BY-NC 4.0);
heads of detected people blurred and the run's timing line removed.}
\label{fig:frames}
\end{figure}

\begin{table}[t]
\centering\small
\begin{tabular}{@{}llrrrrll@{}}
\toprule
Ahead & Distance & $n$ (people) & final & Tier A & CV & final $-$ CV & final $-$ Tier A \\
\midrule
1.2\,s & all & 1,870 (337) & 2.18 & 2.96 & 2.35 & $-0.17$ {\scriptsize[$-0.27$, $-0.06$]} & $-0.79$ {\scriptsize[$-0.97$, $-0.59$]} \\
 & <8 m & 1,053 (185) & 1.24 & 2.36 & 1.22 & $+0.01$ {\scriptsize[$-0.13$, $+0.18$]} & $-1.13$ {\scriptsize[$-1.37$, $-0.84$]} \\
 & 8--15 m & 347 (103) & 2.12 & 2.55 & 2.53 & $-0.41$ {\scriptsize[$-0.58$, $-0.25$]} & $-0.43$ {\scriptsize[$-0.60$, $-0.28$]} \\
 & >15 m & 470 (107) & 4.33 & 4.62 & 4.75 & $-0.42$ {\scriptsize[$-0.63$, $-0.24$]} & $-0.29$ {\scriptsize[$-0.44$, $-0.14$]} \\
\addlinespace
2.4\,s & all & 1,189 (182) & 3.18 & 4.10 & 3.40 & $-0.23$ {\scriptsize[$-0.40$, $-0.04$]} & $-0.93$ {\scriptsize[$-1.20$, $-0.64$]} \\
 & <8 m & 686 (96) & 2.31 & 3.58 & 2.36 & $-0.06$ {\scriptsize[$-0.30$, $+0.28$]} & $-1.27$ {\scriptsize[$-1.61$, $-0.82$]} \\
 & 8--15 m & 204 (54) & 3.36 & 3.77 & 3.68 & $-0.32$ {\scriptsize[$-0.58$, $-0.08$]} & $-0.41$ {\scriptsize[$-0.61$, $-0.22$]} \\
 & >15 m & 299 (61) & 5.04 & 5.52 & 5.59 & $-0.55$ {\scriptsize[$-0.87$, $-0.24$]} & $-0.48$ {\scriptsize[$-0.73$, $-0.24$]} \\
\addlinespace
4.8\,s & all & 566 (70) & 4.40 & 5.23 & 4.87 & $-0.47$ {\scriptsize[$-0.75$, $-0.11$]} & $-0.83$ {\scriptsize[$-1.18$, $-0.36$]} \\
 & <8 m & 363 (39) & 3.11 & 4.22 & 3.56 & $-0.44$ {\scriptsize[$-0.86$, $+0.19$]} & $-1.10$ {\scriptsize[$-1.57$, $-0.39$]} \\
 & 8--15 m & 62 (19) & 6.42 & 6.22 & 6.36 & $+0.07$ {\scriptsize[$-0.33$, $+0.40$]} & $+0.20$ {\scriptsize[$-0.25$, $+0.67$]} \\
 & >15 m & 141 (22) & 6.83 & 7.40 & 7.60 & $-0.77$ {\scriptsize[$-1.04$, $-0.46$]} & $-0.57$ {\scriptsize[$-0.85$, $-0.25$]} \\
\bottomrule
\end{tabular}

\caption{Held-out clips (advio-01--05, 20, 21), pooled, as in
\cref{tab:phone_dev}. The pre-registered primary comparison is the first
column of differences on the ``all'' rows.}
\label{tab:phone_heldout}
\end{table}

\Cref{tab:phone_heldout} reports the pre-registered evaluation on the seven
held-out clips: \HeldNOne{} forecasts scored at 1.2\,s, made for
\HeldPeopleOne{} tracked people. The primary hypothesis holds: the final
forecaster's NLL against the raw measurement is lower than the CV baseline's
at all three horizons, by \HeldGainOne{} \HeldGainOneCI{}, \HeldGainTwo{}
\HeldGainTwoCI{} and \HeldGainFour{} \HeldGainFourCI{} nats at 1.2, 2.4 and
4.8\,s (95\,\% intervals over people). The gains are, however, less than half
of those on the development clips (\DevGainOne{}, \DevGainTwo{} and
\DevGainFour{}), and they are not uniform. Below, differences are final
$-$ CV, so negative values favour the final forecaster. Beyond 8\,m the
development clips' differences replicate at 1.2\,s (8--15\,m: \HeldMidDiffOne{}
\HeldMidDiffOneCI{}, against \DevMidDiffOne{} on the development clips; beyond
15\,m: \HeldFarDiffOne{} \HeldFarDiffOneCI{}, against \DevFarDiffOne{}), but
within 8\,m, where \HeldNearShareOne{} of the held-out forecasts lie
(\DevNearShareOne{} on the development clips), the difference is
\HeldNearDiffOne{} \HeldNearDiffOneCI{} at 1.2\,s, \HeldNearDiffTwo{}
\HeldNearDiffTwoCI{} at 2.4\,s and \HeldNearDiffFour{} \HeldNearDiffFourCI{} at
4.8\,s, none significant; \HeldNearShareFourFive{} of these near forecasts come
from two clips, advio-04 and advio-05, and advio-04 alone supplies
\NearAdvioFourShareOne{}, \NearAdvioFourShareTwo{} and \NearAdvioFourShareFour{}
of them at 1.2, 2.4 and 4.8\,s. That clip rides escalators for most of its
window with the fallback camera height and ADVIO's reference poses, and the
floor rule's lag compresses its distances by about half during the rides
(\cref{sec:discussion}), so this band mixes distance with clip and geometry.
The smaller pooled gain thus comes mostly from the near band, which two clips
with known geometric problems dominate, combined with a different mix of
distances. At 4.8\,s the pattern changes: 8--15\,m shows no gain
(\HeldMidDiffFour{} \HeldMidDiffFourCI{}), and the pooled difference comes
from beyond 15\,m (\HeldFarDiffFour{} \HeldFarDiffFourCI{}) and from the near
band's non-significant point estimate. Per clip (\cref{tab:perclip_heldout}), the
final forecaster beats CV at 1.2\,s on \HeldClipsBetterOne{} of the seven
clips, significantly on \HeldClipsSigOne{}, and is significantly worse on
advio-05 (\HeldClipFiveDiffOne{} \HeldClipFiveDiffOneCI{}). There the phone
looks down from a gallery onto people one floor below and passes shop windows
whose mannequins are detected and tracked as people (\cref{fig:frames}). The
first violates the assumption of a single walkable floor; the second adds
static non-pedestrians, which a constant-velocity Gaussian with a velocity near
zero predicts almost perfectly, whereas a network trained on walkers spreads
probability ahead; we cannot separate these two causes here, and the
exploratory re-run of \cref{sec:exp_explore} shows that the clip's reference
poses contribute as well.

\begin{table}[t]
\centering\small
\begin{tabular}{@{}lllll@{}}
\toprule
Comparison (raw truth, all distances) & $n$ @1.2\,s & 1.2\,s & 2.4\,s & 4.8\,s \\
\midrule
final $-$ CV (primary) & 1,870 & $-0.17$ {\scriptsize[$-0.27$, $-0.06$]} & $-0.23$ {\scriptsize[$-0.40$, $-0.04$]} & $-0.47$ {\scriptsize[$-0.75$, $-0.11$]} \\
final $-$ Tier A & 1,870 & $-0.79$ {\scriptsize[$-0.97$, $-0.59$]} & $-0.93$ {\scriptsize[$-1.20$, $-0.64$]} & $-0.83$ {\scriptsize[$-1.18$, $-0.36$]} \\
final $-$ final without segmentation & 1,870 & $+0.06$ {\scriptsize[$-0.02$, $+0.15$]} & $+0.10$ {\scriptsize[$-0.04$, $+0.26$]} & $+0.07$ {\scriptsize[$-0.16$, $+0.36$]} \\
Tier B goals + physics $-$ CV & 1,870 & $+0.58$ {\scriptsize[$+0.41$, $+0.74$]} & $+0.70$ {\scriptsize[$+0.46$, $+0.92$]} & $+0.49$ {\scriptsize[$+0.23$, $+0.72$]} \\
Tier A $-$ CV & 1,870 & $+0.61$ {\scriptsize[$+0.45$, $+0.76$]} & $+0.70$ {\scriptsize[$+0.50$, $+0.88$]} & $+0.36$ {\scriptsize[$+0.12$, $+0.59$]} \\
final $-$ CV, smoothed truth & 2,249 & $-0.16$ {\scriptsize[$-0.27$, $-0.04$]} & $-0.13$ {\scriptsize[$-0.34$, $+0.08$]} & $-0.32$ {\scriptsize[$-0.63$, $+0.09$]} \\
height fixed at 1.47\,m: final $-$ CV & 1,891 & $-0.15$ {\scriptsize[$-0.26$, $-0.03$]} & $-0.22$ {\scriptsize[$-0.40$, $-0.02$]} & $-0.47$ {\scriptsize[$-0.78$, $-0.06$]} \\
floor fixed at first frame: final $-$ CV & 1,169 & $-0.17$ {\scriptsize[$-0.31$, $-0.04$]} & $-0.25$ {\scriptsize[$-0.46$, $-0.03$]} & $-0.51$ {\scriptsize[$-0.78$, $-0.22$]} \\
\bottomrule
\end{tabular}

\caption{Held-out clips: primary, secondary and sensitivity comparisons (raw
truth, all distances, 95\,\% intervals from resampling people; negative
favours the first method). Rows 1--3 and 6--8 are specified by the protocol;
rows 4 and 5 compare pre-registered variants with CV, which the protocol did
not specify. The last two rows change the geometry, so each compares a variant
with its own CV baseline on its own scored forecasts.}
\label{tab:heldout_secondary}
\end{table}

The secondary runs (\cref{tab:heldout_secondary}) add three observations.
First, the physics prior does not transfer: the final forecaster is better
than Tier A by \HeldGainAOne{}, \HeldGainATwo{} and \HeldGainAFour{} nats, and
Tier A, which was marginally better than CV on the development clips, is worse
than CV on the held-out clips (by \HeldTierACVNegOne{} nats at 1.2\,s; a
comparison the protocol did not specify), mostly on advio-04, an escalator clip
whose many near-range forecasts dominate the pool (\cref{tab:perclip_heldout}).
Second, the map decision taken on the development clips did not replicate:
with segmentation in the map the held-out NLL changes by \HeldMapDiffOne{}
\HeldMapDiffOneCI{}, \HeldMapDiffTwo{} and \HeldMapDiffFour{} nats, so the
variant without it would have done slightly, though not significantly, better.
Third, the network's goals pushed through the physics rollout (``Tier B
goals'') are also worse than CV here (\HeldGoalsCVNegOne{} nats at 1.2\,s; not
specified either): on new real video, only the network's own heat map beats
the baseline.

The comparison is insensitive to the camera height used: with the height fixed
at \DevHeightMedian{} (the median of the development clips' estimates) for
every clip instead of each clip's estimate, the final forecaster's advantage
over CV is \HeldSensOne{}, \HeldSensTwo{} and \HeldSensFour{} nats
(\HeldNSens{} forecasts; the clips' horizon corrections were kept). This does
not show that the heights are right: a wrong height rescales the input and the
target alike and is largely invisible to this score (\cref{sec:discussion}).
Against the smoothed truth, which gave the larger advantage on the development
clips, the advantage is \HeldSmoothOne{} \HeldSmoothOneCI{} at 1.2\,s but not
significant at 2.4\,s (\HeldSmoothTwo{} \HeldSmoothTwoCI{}) and 4.8\,s
(\HeldSmoothFour{} \HeldSmoothFourCI{}). The floor fix matters on the clips
that change level (\cref{tab:floor_heldout}): with the floor held at the first
frame, advio-04, which descends \EscalatorFourGTAbs{} on escalators, yields
\FloorFourFixedN{} scored forecasts instead of \FloorFourFollowN{}, with no
track for \FloorFourFixedNoTrack{} of the clip; on advio-01, which rises
\EscalatorOneGTAbs{} on escalators, no forecast is made for anyone within 8\,m
(\FloorOneNearFixed{} instead of \FloorOneNearFollow{} of the forecasts). Over
all seven clips, \HeldNFixed{} instead of \HeldNOne{} forecasts are scored.
The coverage of the final forecaster's regions on these clips is discussed in
\cref{sec:discussion} (\cref{tab:coverage}).

\subsection{Exploratory analyses (not pre-registered)}
\label{sec:exp_explore}
The analyses in this section were not specified in the protocol; we ran them
after its analysis was complete and after seeing its results, partly prompted
by an internal review of the draft. Most use re-runs of the final
configuration on all twelve clips that store a few more quantities per
forecast (\RecalMismatchText{} the stored scores differed from those of the
original runs). All intervals resample people (tracks) unless stated
otherwise.

\paragraph{Pose source.} On three held-out clips (02, 04, 05) the converter
used ADVIO's reference poses because ARKit failed its whole-sequence
plausibility check; they carry \PoseRefShareHeld{} of the held-out forecasts.
The final forecaster's difference to CV is similar on these clips and on the
four ARKit clips at 1.2\,s (\PoseReferenceOne{} \PoseReferenceOneCI{} against
\PoseArkitOne{} \PoseArkitOneCI{}); at 4.8\,s it is smaller and not
significant on the reference-pose clips (\PoseReferenceFour{}
\PoseReferenceFourCI{} against \PoseArkitFour{} \PoseArkitFourCI{}). On
advio-02 and advio-05 ARKit fails the check only outside the 90\,s windows
(\cref{tab:clips}), so we re-converted these two clips with ARKit's poses
(same windows; camera height and horizon re-estimated from the people) and
ran the final configuration once on each. This changes the poses, the camera
heights, the horizon corrections and which forecasts are scored, not the poses
alone. With ARKit's poses the people-based camera heights become plausible
(\ArkitHeightTwo{} and \ArkitHeightFive{}, with horizon corrections of
\ArkitPitchTwo{} and \ArkitPitchFive{}, against \RefPeopleHeightTwo{},
rejected, and \RefPeopleHeightFive{} with the reference poses), many more
forecasts are scored (\TwoFiveArkitNOne{} instead of \TwoFiveReferenceNOne{} at
1.2\,s on the two clips), and the distances look different (on advio-02,
\TwoFarShareArkit{} instead of \TwoFarShareReference{} of the forecasts are for
people beyond 15\,m; on advio-05, \FiveNearShareArkit{} instead of
\FiveNearShareReference{} within 8\,m). Per clip, the final forecaster's
difference to CV at 1.2, 2.4 and 4.8\,s changes on advio-02 from
\ClipTwoReferenceOne{}, \ClipTwoReferenceTwo{}, \ClipTwoReferenceFour{} to
\ClipTwoArkitOne{}, \ClipTwoArkitTwo{}, \ClipTwoArkitFour{}, and on advio-05
from \ClipFiveReferenceOne{}, \ClipFiveReferenceTwo{}, \ClipFiveReferenceFour{}
to \ClipFiveArkitOne{}, \ClipFiveArkitTwo{}, \ClipFiveArkitFour{} (pooled over
the two clips: \TwoFiveArkitOne{} \TwoFiveArkitOneCI{} at 1.2\,s, against
\TwoFiveReferenceOne{} with the reference poses). The implausible camera
heights, and much of the failure on advio-05, thus seem to come from the
reference poses' attitude rather than from the scene alone (the lower floor
and the mannequins are still there); with the phone's own poses these two
clips favour the network more, not less.

\paragraph{Clips as the unit.} The intervals above treat the seven clips as
fixed. Resampling clips instead of people gives \ClipBootOneCI{},
\ClipBootTwoCI{} and \ClipBootFourCI{} at 1.2, 2.4 and 4.8\,s: only the
4.8\,s difference excludes zero, and even there only \HeldClipsBetterFour{}
of the seven clips favour the network and advio-04 holds \AdvioFourShareFour{}
of the forecasts; with seven unequal clusters the percentile bootstrap is
optimistic (an exact sign-flip test over the seven clip means gives one-sided
$p = \SignFlipOne{}$, $\SignFlipTwo{}$ and $\SignFlipFour{}$). The 1.2\,s mean
rests largely on advio-02: without it, the gain is \LooTwoGainOne{} nats.

\paragraph{The tails.} The score caps a miss at 10.6~nats. At 1.2\,s,
\FloorCVNOne{} of the CV baseline's scores (\FloorCVShareOne{}) reach this
floor, \FloorCVAdvioTwoOne{} of them on advio-02, whose people are mostly
beyond 15\,m with the reference poses (\TwoFarShareReference{}; with ARKit's
poses \TwoFarShareArkit{}), against \FloorOursNOne{} of the final
forecaster's (\FloorOursShareOne{}). Forecasts on which either reaches the
floor contribute \FloorContribOne{} of the \HeldGainOne{}-nat mean gain;
without them the difference is \NoFloorOne{} \NoFloorOneCI{}. The median
difference, which the floor does not affect, is \MedianOne{} \MedianOneCI{},
so typical forecasts do improve. At 4.8\,s, \FloorCVShareFour{} of the
baseline's and \FloorOursShareFour{} of the final forecaster's scores are at
the floor, and \OutsideGridFour{} of the raw measurements lie outside the
25.6\,m grid. Part of the measured gain is therefore a matter of how each
forecaster's tails handle rare, large measurement and tracking errors.

\paragraph{Recalibrating both forecasters.} The protocol's CV baseline takes
its widths from benchmark training data, which need not suit phone video; a
fair competitor could be tuned on phone data, and so could our forecaster. We
recalibrated both on the development clips, per horizon and distance band, in
two ways (\cref{tab:recal}; fitted values and development sample sizes in
\cref{tab:recal_fits}; the 4.8\,s near band rests on only \DevNNearFour{}
development forecasts). With one width, the baseline's width is refitted by maximum
likelihood (for people beyond 15\,m at 4.8\,s it becomes \RecalSigmaFarFour{},
against \ProtocolSigmaFarFour{} on average in the protocol's baseline), and our
forecast receives the extra isotropic blur (0, 0.25, 0.5, 1, 2 or 4\,m) that
minimises its NLL there; the blur can only widen our forecast, and zero blur
was chosen in \BlurZeroBands{} of the nine bands. With mixtures, the baseline becomes a
mixture of two Gaussians with a common mean (two widths and a weight), and our
forecast, after its best blur, is mixed with a Gaussian around the
constant-velocity mean (a weight and a width); within 15\,m the fitted
component is narrow (at most \OursMixWidthMax{} wider than the measurement
noise) and carries
\OursMixWeightRange{} of the mass, so the recalibrated
forecaster there is partly a sharp constant-velocity forecaster, while beyond
15\,m it is broad. Recalibration helps the baseline more than our
forecaster, much more at 2.4 and 4.8\,s: with mixtures, the held-out NLL of
the baseline falls by \MixCVGainOne{}, \MixCVGainTwo{} and \MixCVGainFour{}
nats at 1.2, 2.4 and 4.8\,s, ours by \MixOursGainOne{}, \MixOursGainTwo{} and
\MixOursGainFour{}. The baseline's two-Gaussian mixture is even better than
the final forecaster as configured at every horizon (\OursVsMixCVOne{}
\OursVsMixCVOneCI{} at 1.2\,s, \OursVsMixCVFour{} \OursVsMixCVFourCI{} at
4.8\,s; over clips \OursVsMixCVOneClipCI{} and \OursVsMixCVFourClipCI{}).
After both are recalibrated, the final forecaster remains better at 1.2\,s
(\RecalRecalVsRefitOne{} \RecalRecalVsRefitOneCI{} with one width, \MixOne{}
\MixOneCI{} with mixtures), not significantly at 2.4\,s
(\RecalRecalVsRefitTwo{} \RecalRecalVsRefitTwoCI{}; \MixTwo{} \MixTwoCI{}),
and at 4.8\,s the point estimates favour the recalibrated baseline
(\RecalRecalVsRefitFour{} \RecalRecalVsRefitFourCI{}; \MixFour{}
\MixFourCI{}). When clips are resampled, even the 1.2\,s intervals include
zero (\RecalRecalVsRefitOneClipCI{} and \MixOneClipCI{}). Split by distance
(post hoc), the forecasters recalibrated with mixtures differ: beyond 8\,m the
network is
better at 1.2\,s (8--15\,m: \MixMidOne{} \MixMidOneCI{}; beyond 15\,m:
\MixFarOne{} \MixFarOneCI{}) and at 2.4\,s (\MixMidTwo{} \MixMidTwoCI{};
\MixFarTwo{} \MixFarTwoCI{}); within 8\,m it is not better at 1.2\,s
(\MixNearOne{} \MixNearOneCI{}) and worse at 2.4\,s (\MixNearTwo{}
\MixNearTwoCI{}), so the pooled 2.4\,s difference is near zero. Against the
smoothed truth, the baseline with a refitted single width matches or, by point
estimate, beats the unrecalibrated final forecaster at every horizon
(\SmoothRefitOne{} \SmoothRefitOneCI{}, \SmoothRefitTwo{} \SmoothRefitTwoCI{}
and \SmoothRefitFour{} \SmoothRefitFourCI{}). The point estimates thus suggest
that beyond 2\,s most of the pre-registered advantage comes from the
benchmark-fitted baseline's widths and tails being wrong for phone video;
whether a more flexible recalibration of our forecaster would recover an
advantage there is untested. To score the recalibrated baselines exactly like
the other forecasts, on each forecast's own grid, we reconstructed these grids
from the stored rows: re-scoring the protocol's CV in this way reproduces its
stored raw-truth scores to within 0.001~nats for \RecalGridAgree{} of the
forecasts (mean difference \RecalGridMeanDiff{} nats; for the smoothed truth,
\SmoothGridAgree{}).

\begin{table}[t]
\centering\small\setlength{\tabcolsep}{4.5pt}
\begin{tabular}{@{}lrrrrrrrr@{}}
\toprule
 & \multicolumn{2}{c}{as configured} & \multicolumn{3}{c}{one width (CV) / blur (final)} & \multicolumn{3}{c}{mixtures} \\
\cmidrule(lr){2-3}\cmidrule(lr){4-6}\cmidrule(lr){7-9}
Ahead & final & CV & final & CV & final $-$ CV & final & CV & final $-$ CV \\
\midrule
1.2\,s & 2.18 & 2.35 & 2.18 & 2.36 & \makecell[r]{$-0.18$\\[-1pt]{\scriptsize[$-0.28$,\,$-0.07$]}} & 1.91 & 2.05 & \makecell[r]{$-0.14$\\[-1pt]{\scriptsize[$-0.20$,\,$-0.07$]}} \\
2.4\,s & 3.18 & 3.40 & 3.17 & 3.27 & \makecell[r]{$-0.10$\\[-1pt]{\scriptsize[$-0.26$,\,$+0.07$]}} & 2.87 & 2.90 & \makecell[r]{$-0.04$\\[-1pt]{\scriptsize[$-0.15$,\,$+0.07$]}} \\
4.8\,s & 4.40 & 4.87 & 4.36 & 4.29 & \makecell[r]{$+0.07$\\[-1pt]{\scriptsize[$-0.21$,\,$+0.41$]}} & 3.98 & 3.93 & \makecell[r]{$+0.05$\\[-1pt]{\scriptsize[$-0.10$,\,$+0.21$]}} \\
\bottomrule
\end{tabular}

\caption{Exploratory (not pre-registered): held-out clips, raw truth, mean
NLL (nats) of the final forecaster and the CV baseline as configured and after
recalibrating both on the development clips (per horizon and distance band),
with one width (our forecast: an extra blur) or with mixtures (CV: two
Gaussians with a common mean; our forecast: mixed with a Gaussian around the
constant-velocity mean), and the paired difference between the recalibrated
forecasters with a 95\,\% interval from resampling people. All are scored on
each forecast's own grid; the fitted values are in \cref{tab:recal_fits}.}
\label{tab:recal}
\end{table}

\subsection{Runtime}
\label{sec:exp_runtime}

\begin{table}[t]
\centering\small
\begin{tabular}{@{}lrrrrrr@{}}
\toprule
 & detection + tracking & map & forecast & drawing & total & frames/s \\
\midrule
GPU (RTX 4000 Ada) + CPU & 20 & 14 & 15 & 34 & 83 & 12.0 \\
CPU only (Xeon w7-2495X) & 135 & 26 & 36 & 34 & 231 & 4.3 \\
\bottomrule
\end{tabular}

\caption{Mean processing time per frame (ms) of the final configuration
(segmentation in the map) over the first 30\,s (\LatFrames{} frames,
720$\times$1280) of the advio-06 window, a crowded mall clip, with no other
experiment running; all frames are included and reading the video is not.
``Detection + tracking'' also includes lifting and the floor update; ``map'' is
every map update, including segmentation at 2\,Hz; ``forecast'' is at most one
person's forecast per frame, including the preparation of its display layer.
The detector runs on ONNX Runtime (CUDA) in the GPU configuration and on
OpenCV's DNN module in the CPU one; the evaluation runs used the latter. Both
configurations tracked the same number of people throughout.}
\label{tab:latency}
\end{table}

With detection, segmentation and the network on the GPU (\cref{tab:latency}),
the pipeline processes \LatGpuFps{} frames per second (\LatGpuTotal{}\,ms per
frame), of which detection and tracking take \LatGpuTrack{}\,ms, map updates
\LatGpuMap{}\,ms, forecasting \LatGpuForecast{}\,ms and drawing \LatGpuRender{}\,ms;
on the CPU alone it processes \LatCpuFps{} frames per second, most of the time
going to detection (\LatCpuTrack{}\,ms). Neither keeps up with the 30 frames
per second of our converted clips (ADVIO records at 60); without drawing, the
GPU configuration would process about \LatGpuNoDrawFps{} per second. Because at most one person is re-forecast per frame,
the processing rate also bounds how often forecasts are refreshed in a crowd:
in our offline runs every frame was processed, allowing up to 30 forecasts per
second of video, but a live system at \LatGpuFps{} frames per second could
refresh at most that many. These are workstation timings; we have not
implemented the pipeline on a phone.

\section{Discussion and limitations}
\label{sec:discussion}

\paragraph{What the phone-video numbers mean.} The self-consistency scores
compare forecasts with the system's own later measurements. Lifting errors
that persist over seconds (a wrong camera height, a slope ahead, a
mis-estimated horizon) affect the input history and the target in the same
way and are invisible to this score; identity switches and survivorship
(only people still tracked at the horizon are scored) bias it in ways that
need not favour either method, and the tracker's constant-velocity gate
decides which measurements count as the same person. The raw-truth score
removes the smoother's constant-velocity bias but adds the measurement noise
of \cref{eq:noise}, whose scale was set on a development clip, and rare large
errors reach the score's floor, so the means depend on how each forecaster's
tails treat them (\cref{sec:exp_explore}). The comparison between forecasters
is paired, because they are scored on identical forecasts and targets; the
absolute values are not comparable with the benchmarks, and none of this
replaces an evaluation against independent ground truth, such as a phone
recording with a synchronised overhead camera or LiDAR, which we could not find
publicly.

\paragraph{Calibration.} On the benchmarks the final forecaster is
under-confident: its 50\,\% regions contain the truth \BenchSddHeatCovFiftyOne{},
\BenchSddHeatCovFiftyTwo{} and \BenchSddHeatCovFiftyFour{} of the time on SDD
at 1.2, 2.4 and 4.8\,s and \BenchEgoHeatCovFiftyOne{},
\BenchEgoHeatCovFiftyTwo{} and \BenchEgoHeatCovFiftyFour{} on EgoTraj-Bench,
and its 90\,\% regions \BenchSddHeatCovNinetyOne{}, \BenchSddHeatCovNinetyTwo{},
\BenchSddHeatCovNinetyFour{} and \BenchEgoHeatCovNinetyOne{},
\BenchEgoHeatCovNinetyTwo{}, \BenchEgoHeatCovNinetyFour{}
(\cref{fig:calibration}). The softened training target (0.3\,m) and the hedge
both add width; without the hedge, the network's 50\,\% regions are closer to
nominal but its 90\,\% regions are too narrow at 4.8\,s
(\BenchSddNetAloneCovNinetyFour{} on SDD, \BenchEgoNetAloneCovNinetyFour{} on
EgoTraj-Bench). On held-out phone video, the cores of its maps are about right
at 1.2\,s and too narrow beyond: its 50\,\% regions contain the raw
measurement \HeldCovFiftyOne{}, \HeldCovFiftyTwo{} and \HeldCovFiftyFour{} of
the time at 1.2, 2.4 and 4.8\,s (CV: \HeldCovFiftyCVOne{},
\HeldCovFiftyCVTwo{}, \HeldCovFiftyCVFour{}), its
80\,\% regions \HeldCovEightyOne{}, \HeldCovEightyTwo{}, \HeldCovEightyFour{}
(CV: \HeldCovEightyCVOne{}, \HeldCovEightyCVTwo{}, \HeldCovEightyCVFour{}),
and its 90\,\% regions \HeldCovNinetyOne{}, \HeldCovNinetyTwo{},
\HeldCovNinetyFour{} (CV: \HeldCovNinetyCVOne{}, \HeldCovNinetyCVTwo{},
\HeldCovNinetyCVFour{}; \cref{tab:coverage}); for people beyond 15\,m its
90\,\% regions contain the measurement only \HeldCovNinetyFarFour{} of the time
at 4.8\,s (CV: \HeldCovNinetyCVFarFour{}). The CV baseline's single Gaussian,
in turn, has too little mass in the tails (\cref{sec:exp_explore}). The hedge weights were fitted for Tier A
and not re-tuned for the network, and the final configuration was not
recalibrated on phone data; the exploratory recalibration of
\cref{sec:exp_explore} shows that this matters for the comparison. A
principled recalibration on phone data, for instance
conformal~\citep{lindemann2023safe,dixit2023adaptive}, is the obvious next
step, with the caveat that its target would again be the system's own
measurements.

\begin{table}[t]
\centering\small
\begin{tabular}{@{}llrrrrrr@{}}
\toprule
 & & \multicolumn{3}{c}{final} & \multicolumn{3}{c}{CV} \\
\cmidrule(lr){3-5}\cmidrule(lr){6-8}
Clips & Ahead & 50\,\% & 80\,\% & 90\,\% & 50\,\% & 80\,\% & 90\,\% \\
\midrule
development & 1.2\,s & 0.65 & 0.84 & 0.89 & 0.60 & 0.78 & 0.84 \\
 & 2.4\,s & 0.51 & 0.74 & 0.83 & 0.47 & 0.65 & 0.71 \\
 & 4.8\,s & 0.39 & 0.63 & 0.73 & 0.40 & 0.54 & 0.61 \\
\addlinespace
held-out & 1.2\,s & 0.54 & 0.75 & 0.86 & 0.64 & 0.80 & 0.86 \\
 & 2.4\,s & 0.41 & 0.65 & 0.79 & 0.62 & 0.75 & 0.80 \\
 & 4.8\,s & 0.33 & 0.58 & 0.76 & 0.66 & 0.73 & 0.77 \\
\bottomrule
\end{tabular}

\caption{Phone clips, raw truth, all distances: share of the raw measurements
inside the 50, 80 and 90\,\% highest-density regions of the final forecaster
and of the CV baseline (calibrated forecasts reach the nominal level).}
\label{tab:coverage}
\end{table}

\paragraph{Geometry without depth.} The floor rule of \cref{sec:floor}
handles drift of the phone's altitude and the user's own change of level, but
not terrain that slopes between the user and the person, multi-level spaces in
which people stand on another floor (as on advio-05), or a wrong camera
height. While the phone's altitude changes at a steady rate $v$, as on an
escalator, the filtered altitude lags by about $v\tau$, which inflates or
compresses distances by $(h \pm v\tau)/h$ until the ride ends. On advio-04,
which spends \RideShareFour{} of its window on escalators
(\RideForecastShareFour{} of its forecasts are made then) at
\RideSpeedFour{}\,m\,s$^{-1}$ of vertical speed on average according to the
reference poses, the lag is about \RideLagFour{}, which compresses distances by
about \RideScaleFour{} during the rides (exploratory estimates). The VIO's own
vertical velocity could remove this lag. The people-based camera height was
rejected as implausible on two of the seven held-out clips (02, 04) and is
implausibly high for a handheld phone on two more (01 and 05, above 1.8\,m),
although those pass the accepted range of 1.0--1.9\,m, which is therefore
generous (\cref{tab:clips}); with ARKit's instead of the reference poses, the
estimates on advio-02 and advio-05 become plausible (\cref{sec:exp_explore}),
so the attitude of the poses matters as much as the people. The measurement
noise of \cref{eq:noise} is capped at 2\,m, which it reaches at about 28\,m,
while rays are kept to 60\,m; \BeyondTwentyEightHeld{} of the held-out
forecasts are made for people beyond 28\,m, whose noise is therefore
under-stated. ADVIO's reference poses are
themselves level on sloping streets, so they cannot serve as a reference for
vertical motion there; we relied on the barometer for that comparison. The
camera height was estimated per clip from all people in the window, offline;
a live system would estimate it online (implemented, not evaluated). On
phones with LiDAR, positions within about 5\,m would be measured directly; the
code supports this, but no public recording of pedestrians from a LiDAR phone
was available, so it is untested on real data.

\paragraph{Accuracy versus calibration.} Best-of-20 errors of our sample-based
variants are far from the state of the art on both benchmarks
(\cref{sec:exp_bench}); the final heat-map forecaster produces no trajectories
at all. Our claims concern densities, their calibration and their use on real
phone video, not displacement accuracy. A modern trajectory generator could
replace the heat-map network inside the same pipeline if it provided
densities.

\paragraph{Runtime.} On a workstation the pipeline processes \LatGpuFps{}
frames per second with a GPU and \LatCpuFps{} on the CPU alone
(\cref{sec:exp_runtime}), short of the 30 of our converted clips. A phone has
far less compute, so an on-device version would need a lighter detector,
segmentation at a lower rate or not at all (it did not improve the held-out
scores), and drawing on the phone's GPU. The forecaster itself is cheap: one
evaluation of a 1\,M-parameter network per person at up to 2.5\,Hz.

\paragraph{Data and ethics.} The recordings show members of the public in a
mall, a metro station and streets; we use ADVIO under its CC BY-NC 4.0 licence,
publish no images beyond a few illustrative frames, in which we blurred the
heads of detected people, and do not attempt to identify anyone. A forecaster like this one must
not be relied upon for the safety of a blind or low-vision user without a
dedicated evaluation with such users; on held-out real video its 90\,\%
regions miss the measured position \HeldMissNinetyRange{} of the time,
depending on the horizon.

\section{Conclusion}
\label{sec:conclusion}
We built a pipeline that needs only a phone's camera and VIO pose (in this
evaluation, ADVIO's reference poses replaced ARKit's on four of the twelve
clips) and forecasts pedestrians as probability maps in a gravity-aligned
world frame, and evaluated its densities by likelihood and coverage on two
benchmarks and,
without ground truth, on handheld video under a protocol that was fixed before
the test clips were run. Three lessons are likely to outlast this particular
system. First, monocular ground-plane lifting from a moving phone should not
hold a floor plane fixed in the world, because the VIO altitude drifts and
users change level; otherwise positions are silently rescaled. Keeping the
camera's height above the floor largely avoids this, but lags by about
$v\tau$ during sustained vertical motion such as an escalator ride. Second, a
heat-map network trained with a negative log-likelihood loss gives much
better densities than a physics prior, on the benchmarks and on real video,
where the physics prior did not even beat constant velocity on new clips. On
real phone video, however, its advantage over a constant-velocity baseline
with benchmark-fitted widths was less than half as large on new clips as on
the development clips, absent in the nearest distance band (which one
escalator clip dominates), and reversed on a clip evaluated with ADVIO's
reference poses, where a re-run with the phone's own poses reverses it again.
In exploratory analyses it is fragile: it depends on a few clips and on the
baseline's light tails, a constant-velocity baseline refitted on phone data
as a two-Gaussian mixture beats the final forecaster as configured, and after
both are recalibrated the pooled advantage remains only at 1.2\,s (over
people, not over clips); split by distance, it is confined to people beyond
8\,m. Third, evaluation on real video without ground truth can be made
informative --- scores against raw later measurements with noise-matched
baselines, intervals that resample clips as well as people (with few clips
both are approximate), an analysis fixed in advance --- but it measures
consistency with one's own tracker, not the truth, and a baseline calibrated
on the same kind of data, with realistic tails, is the comparison that
matters. A phone recording with independent ground truth for the pedestrians,
phones with LiDAR, recalibration on phone data and an on-device
implementation are the next steps.

\bibliographystyle{plainnat}
\bibliography{refs}

\clearpage
\appendix
\section{Training details of the heat-map network}
\label{app:training}

\paragraph{Data.} SDD: the pedestrian windows of the TrajNet training split
(8 observed and 12 future positions at 2.5\,Hz), in metres via OpenTraj's
per-video scales, with Y-Net's hand-labelled masks (pavement, road,
structure, terrain, tree; unlabelled cells are \emph{unknown}) downsampled to
0.2\,m by majority vote; five scenes are held out for validation. EgoTraj-Bench:
the TBD subset, data release v1.1; invalid history steps are linearly
interpolated (held constant at the ends) and windows with fewer than two valid
history steps are skipped; the map is all \emph{unknown}. Neighbours are
represented by their last two positions: on SDD, the other agents observed at
both of the last two observed steps; on EgoTraj-Bench, the other agents of
the window's sequence (with their histories interpolated in the same way). In
total \TrainWindowsSdd{} SDD and \TrainWindowsEgo{} EgoTraj-Bench training
windows.

\paragraph{Input encoding.} The $128\times128$ crop (0.2\,m cells) is
centred horizontally on the person's last position, with the person 38 rows
(30\,\%) from the back edge, and rotated so that the heading (from the
displacement over the last three steps, if longer than 0.15\,m; otherwise a
random direction in training and the world $x$ axis at inference) points
along the rows. Past positions and neighbours are drawn as unit-peak
Gaussians of 1.5 cells. Training crops are cut from a 48\,m map, inference
crops from the 25.6\,m local map (\cref{sec:forecasters}).

\paragraph{Augmentation} (training only). History jitter: Gaussian noise with a
standard deviation drawn uniformly from $[0, 0.12]$\,m per window; with
probability 0.3 the first 1--6 history steps are dropped; with probability 0.05
the earlier part of the history is shifted by a Gaussian of 1\,m (an identity
switch); heading noise of $15^\circ$; with probability 0.6 a visibility mask:
a camera 4--14\,m from the person at a uniformly random bearing, aimed at the
person with a Gaussian error of $12^\circ$, with a field of view of
$60$--$75^\circ$ and a range of 15--28\,m, everything outside of which becomes
\emph{unknown}; left--right mirroring with probability 0.5.

\paragraph{Optimisation.} AdamW~\citep{loshchilov2019decoupled} (learning rate
$10^{-3}$, weight decay $10^{-4}$), one-cycle schedule~\citep{smith2019super}
with 5\,\% warm-up, batch 32, 60 epochs, mixed precision, gradient clipping at
5. After every epoch the network is evaluated on SDD validation and EgoTraj-Bench
validation; the checkpoint with the lowest mean of the two NLLs (each averaged
over 1.2, 2.4 and 4.8\,s) is kept. This selection NLL is computed on the
network's crop, without the map mask and the probability floor and with the
true cell found by truncating its crop coordinates, so it differs slightly from
the benchmark NLL of \cref{sec:benchmarks}. The model used throughout (seed 0)
was trained on an NVIDIA A100 (Google Colab), the two further seeds on an
NVIDIA RTX 4000 Ada; the augmentation draws from a generator seeded by the
clock, and cuDNN autotuning adds further nondeterminism, so runs are not
bit-reproducible and we report several.

\begin{figure}[ht]
\centering
\includegraphics[width=\linewidth]{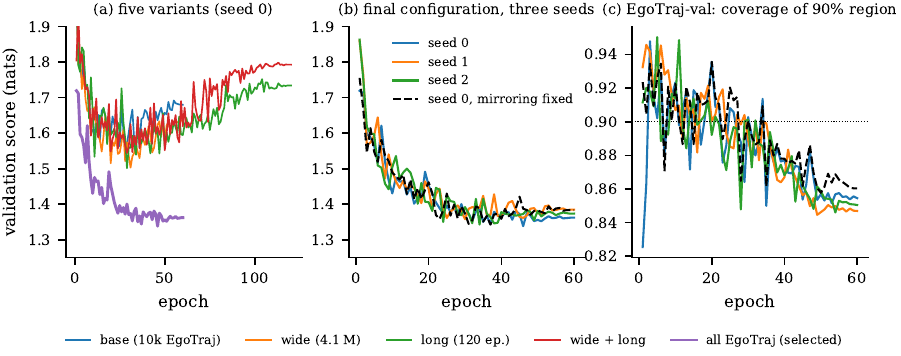}
\caption{Training curves: the checkpoint-selection score (mean of the SDD and
EgoTraj-Bench validation NLLs) of the five variants, the last of which is the
final configuration (a); of the final configuration with three seeds and,
dashed, retrained with the mirroring fix of \cref{app:training} (b); and the
coverage of the 90\,\% region on EgoTraj-Bench validation (c).}
\label{fig:training}
\end{figure}

\paragraph{A bug in the mirroring augmentation.} After all reported runs we
found an off-by-one in the left--right mirroring: the input was flipped
correctly (column $i \mapsto 127-i$), but the target column $c$ became $128-c$
instead of $127-c$, so in mirrored training samples the target lay one cell
(0.2\,m) sideways of where the mirrored input put it (a target in column 0
fell outside the crop and was dropped from the loss). All networks reported
in this paper were trained with this bug, which shifts half of the training
targets by 0.2\,m to the same side of the crop. To measure its effect without
using the test sets, we retrained the final configuration once with the fix
(seed 0, so with the same initial weights as the reported network) and scored
it on the validation sets exactly as for the seeds' choice of the fusion
weight. It chose the same weight ($w = 1$ for both profiles), and its
checkpoint-selection score, \MirrorFixSelect{}, lies within the range of the
three seeds trained with the bug (\MirrorBugSelectMin{}--\MirrorBugSelectMax{};
\cref{fig:training}). The final configuration's mean validation NLL with the
fixed network is \MirrorFixSddVal{} on SDD and \MirrorFixEgoVal{} on
EgoTraj-Bench, against \MirrorBugSddValMin{}--\MirrorBugSddValMax{} and
\MirrorBugEgoValMin{}--\MirrorBugEgoValMax{} with the bug: within the range on
EgoTraj-Bench and slightly above it (worse) on SDD. The fix thus made no
measurable improvement; we did not evaluate the fixed network on the test sets
or on the phone clips.

\paragraph{Model selection history.} Five variants were trained once each
(\cref{tab:sweep}); the last of them is the final configuration, chosen on the
validation score only. Using all EgoTraj-Bench training windows instead of the
first 10,000 improved the score by far the most; four times the parameters or
twice the epochs improved it much less, and both together less still (single
runs).

\begin{table}[ht]
\centering\small
\begin{tabular}{@{}lrrrrrrr@{}}
\toprule
Variant & Params & EgoTraj windows & Epochs & Best epoch & SDD val & EgoTraj val & Score \\
\midrule
base & 1.02\,M & 10,000 & 60 & 36 & 0.995 & 2.126 & 1.561 \\
wide & 4.06\,M & 10,000 & 60 & 31 & 0.970 & 2.036 & 1.503 \\
long & 1.02\,M & 10,000 & 120 & 29 & 1.015 & 1.989 & 1.502 \\
wide + long & 4.06\,M & 10,000 & 120 & 29 & 1.028 & 2.061 & 1.544 \\
all EgoTraj windows & 1.02\,M & all & 60 & 46 & 0.979 & 1.698 & \textbf{1.338} \\
\bottomrule
\end{tabular}

\caption{The five variants trained, the last being the final configuration
(one run each, seed 0; NLL on the validation sets at the best epoch, mean over
1.2, 2.4 and 4.8\,s; the selection score is the mean of the two).}
\label{tab:sweep}
\end{table}

\section{Synthetic check of the floor rule}
\label{app:synthetic}
A synthetic moving-camera world with known pedestrian positions (rendered
frames, ground-truth detections, no LiDAR, camera height known: 1.45\,m)
isolates the floor rule of \cref{sec:floor}; it checks the geometry of
\cref{eq:scale,eq:floor}, not the rule under real detection noise. After 4\,s the reported camera altitude jumps
by $\Delta z$, as a vertical tracking drift would; positions are evaluated
from 12\,s on (four time constants later), \SynthDetections{} detections in
total; the error of a placed detection is its distance to the nearest true
person (a lower bound when placements are far off). With $\Delta z = +1.2$\,m,
the floor held at its first-frame height places people with a median error of
\SynthFixedUpErr{}, the following floor with \SynthFollowUpErr{}. With $\Delta z = -1.6$\,m (more than the camera
height), the fixed floor places \SynthFixedDownPlaced{} of them (no ray meets
it), the following floor all of them, with a median error of
\SynthFollowDownErr{} (\path{scripts/paper_synthetic_floor.py}; the
following floor in this scenario is also a unit test).

\section{Additional results}
\label{app:results}

\begin{table}[ht]
\centering\small\setlength{\tabcolsep}{4pt}
\begin{tabular}{@{}llllrlrlrrrr@{}}
\toprule
Clip & Role & Venue & Crowd & Path & Poses & \multicolumn{3}{c}{Camera height} & \multicolumn{3}{c}{Height change (m)} \\
\cmidrule(lr){7-9}\cmidrule(lr){10-12}
 & & & & (m) & & (m) & source & horizon & ARKit & GT & barom. \\
\midrule
advio-06 & dev & Mall & high & 127 & ARKit & 1.42 & people (283) & $-0.7^\circ$ & $+1.23$ & $-0.04$ & $+0.33$ \\
advio-11 & dev & Metro & high & 98 & ARKit & 1.48 & people (193) & $+0.4^\circ$ & $-0.48$ & $-0.03$ & $-0.10$ \\
advio-12 & dev & Metro & high & 111 & ARKit & 1.47 & people (61) & $0.0^\circ$ & $+0.28$ & $+0.39$ & $+0.12$ \\
advio-22 & dev & Outdoor & high & 106 & ARKit & 1.47 & people (220) & $+0.4^\circ$ & $-1.79$ & $-0.04$ & $-1.54$ \\
advio-23 & dev & Outdoor & high & 98 & ref. & 1.53 & people (48) & $0.0^\circ$ & failed & $-0.01$ & $-4.23$ \\
\midrule
advio-01 & held-out & Mall & moderate & 88 & ARKit & 1.84 & people (140) & $-2.5^\circ$ & $+6.79$ & $+7.78$ & $+7.92$ \\
advio-02 & held-out & Mall & moderate & 102 & ref. & 1.40 & fallback & -- & $+0.50$$^\dagger$ & $+0.01$ & $+0.71$ \\
advio-03 & held-out & Mall & moderate & 96 & ARKit & 1.47 & people (68) & $+1.9^\circ$ & $+0.22$ & $+0.15$ & $-0.16$ \\
advio-04 & held-out & Mall & moderate & 51 & ref. & 1.40 & fallback & -- & failed & $-16.28$ & $-15.35$ \\
advio-05 & held-out & Mall & moderate & 114 & ref. & 1.87 & people (89) & $-1.5^\circ$ & $+0.84$$^\dagger$ & $+0.04$ & $+0.34$ \\
advio-20 & held-out & Outdoor & low & 150 & ARKit & 1.48 & people (97) & $+0.2^\circ$ & $-0.10$ & $-0.01$ & $-0.12$ \\
advio-21 & held-out & Outdoor & low & 134 & ARKit & 1.36 & people (57) & $+0.7^\circ$ & $-0.23$ & $-0.06$ & $+0.19$ \\
\bottomrule
\end{tabular}

\caption{The twelve ADVIO clips (90\,s windows with the most people). Venue
and crowd level from the ADVIO documentation; horizontal path length of the
phone from ADVIO's reference poses; poses used by the pipeline (ARKit unless it
fails a plausibility check over the whole sequence, else ADVIO's reference
poses, ``ref.''; $^\dagger$ARKit plausible within the window but not over the
whole sequence); camera height above the floor, its source (people as the
ruler, with the number of person observations used, or the 1.40\,m fallback
when the estimate was rejected as implausible) and the horizon correction
fitted with it (\cref{sec:floor}); change of the phone's height from the start
to the end of the window according to ARKit, ADVIO's reference poses (GT) and
the phone's barometer.}
\label{tab:clips}
\end{table}

\begin{table}[ht]
\centering\small
\begin{tabular}{@{}lrr@{}}
\toprule
Method & minADE$_{20}$ & minFDE$_{20}$ \\
\midrule
\multicolumn{3}{@{}l}{\emph{SDD TrajNet test, pixels}} \\
S-GAN~\citep{gupta2018social} & 27.23 & 41.44 \\
SoPhie~\citep{sadeghian2019sophie} & 16.27 & 29.38 \\
PECNet~\citep{mangalam2020not} & 9.96 & 15.88 \\
Y-Net~\citep{mangalam2021goals} & 7.85 & 11.85 \\
Tier A (ours) & 13.97 & 21.58 \\
Tier B goals + physics (ours) & 13.54 & 19.71 \\
\midrule
\multicolumn{3}{@{}l}{\emph{EgoTraj-Bench TBD test, metres}} \\
BiFlow~\citep{liu2026egotraj}, released checkpoints, data v1.1 & 0.30 & 0.44 \\
Tier A (ours) & 0.78 & 1.29 \\
Tier B goals + physics (ours) & 0.59 & 0.88 \\
\bottomrule
\end{tabular}

\caption{Best-of-20 errors next to published results (as reported by their
authors; SDD in pixels of the original images, EgoTraj-Bench in metres). Our
values use 20 random samples of the goal-conditioned rollouts; the final
heat-map forecaster produces no samples. On EgoTraj-Bench, only BiFlow has been
reported on the current data release (by its authors, about 0.30/0.44\,m); the
paper's other baselines were run on an earlier release~\citep{liu2026egotraj}.
We evaluate all \EgoTestAgents{} test agents of the release (our loader skips
none).}
\label{tab:published}
\end{table}

\begin{figure}[ht]
\centering
\includegraphics[width=\linewidth]{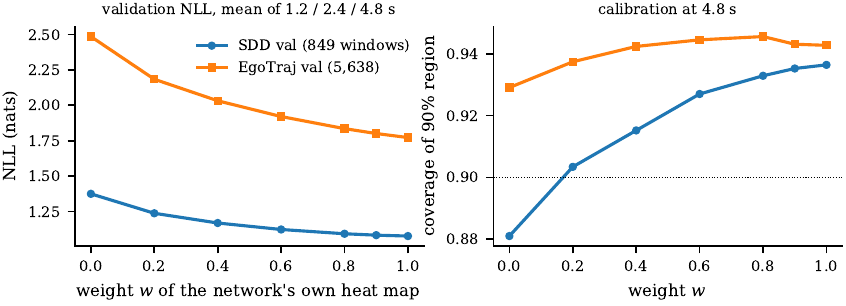}
\caption{Choosing the weight $w$ of the network's own heat map against the
physics rollout steered by the network's goals, on validation data only:
mean NLL over 1.2, 2.4 and 4.8\,s, and coverage of the 90\,\% region at 4.8\,s.}
\label{fig:fusion}
\end{figure}

\begin{table}[ht]
\centering\small
\begin{tabular}{@{}llrrrrr@{}}
\toprule
Clip & poses & $n$ & \multicolumn{3}{c}{final $-$ CV} & Tier A $-$ CV \\
\cmidrule(lr){4-6}\cmidrule(lr){7-7}
 & & & 1.2\,s & 2.4\,s & 4.8\,s & 1.2\,s \\
\midrule
advio-06 & ARKit & 770 & \makecell[r]{$-0.52$\\[-1pt]{\scriptsize[$-0.63$,\,$-0.42$]}} & \makecell[r]{$-1.06$\\[-1pt]{\scriptsize[$-1.23$,\,$-0.86$]}} & \makecell[r]{$-1.32$\\[-1pt]{\scriptsize[$-1.71$,\,$-0.93$]}} & \makecell[r]{$-0.08$\\[-1pt]{\scriptsize[$-0.15$,\,$-0.00$]}} \\
advio-11 & ARKit & 568 & \makecell[r]{$-0.57$\\[-1pt]{\scriptsize[$-0.70$,\,$-0.47$]}} & \makecell[r]{$-1.06$\\[-1pt]{\scriptsize[$-1.28$,\,$-0.86$]}} & \makecell[r]{$-1.63$\\[-1pt]{\scriptsize[$-1.98$,\,$-1.27$]}} & \makecell[r]{$-0.20$\\[-1pt]{\scriptsize[$-0.29$,\,$-0.11$]}} \\
advio-12 & ARKit & 543 & \makecell[r]{$-0.40$\\[-1pt]{\scriptsize[$-0.51$,\,$-0.30$]}} & \makecell[r]{$-0.89$\\[-1pt]{\scriptsize[$-1.12$,\,$-0.69$]}} & \makecell[r]{$-1.09$\\[-1pt]{\scriptsize[$-1.61$,\,$-0.62$]}} & \makecell[r]{$-0.12$\\[-1pt]{\scriptsize[$-0.16$,\,$-0.07$]}} \\
advio-22 & ARKit & 1,107 & \makecell[r]{$-0.36$\\[-1pt]{\scriptsize[$-0.45$,\,$-0.27$]}} & \makecell[r]{$-0.70$\\[-1pt]{\scriptsize[$-0.84$,\,$-0.56$]}} & \makecell[r]{$-1.01$\\[-1pt]{\scriptsize[$-1.28$,\,$-0.74$]}} & \makecell[r]{$-0.04$\\[-1pt]{\scriptsize[$-0.09$,\,$+0.01$]}} \\
advio-23 & reference & 147 & \makecell[r]{$-0.12$\\[-1pt]{\scriptsize[$-0.35$,\,$+0.09$]}} & \makecell[r]{$-0.12$\\[-1pt]{\scriptsize[$-0.66$,\,$+0.37$]}} & \makecell[r]{$-0.19$\\[-1pt]{\scriptsize[$-1.29$,\,$+0.47$]}} & \makecell[r]{$0.00$\\[-1pt]{\scriptsize[$-0.10$,\,$+0.13$]}} \\
\bottomrule
\end{tabular}

\caption{Development clips, per clip: poses used, forecasts scored at 1.2\,s,
and paired raw-truth NLL differences (nats; negative favours the first
method) of the final forecaster to CV at 1.2, 2.4 and 4.8\,s and of Tier A to
CV at 1.2\,s, with 95\,\% intervals from resampling people.}
\label{tab:perclip_dev}
\end{table}

\begin{table}[ht]
\centering\small
\begin{tabular}{@{}llrrrrr@{}}
\toprule
Clip & poses & $n$ & \multicolumn{3}{c}{final $-$ CV} & Tier A $-$ CV \\
\cmidrule(lr){4-6}\cmidrule(lr){7-7}
 & & & 1.2\,s & 2.4\,s & 4.8\,s & 1.2\,s \\
\midrule
advio-01 & ARKit & 130 & \makecell[r]{$-0.30$\\[-1pt]{\scriptsize[$-0.48$,\,$-0.11$]}} & \makecell[r]{$-0.59$\\[-1pt]{\scriptsize[$-0.89$,\,$-0.23$]}} & \makecell[r]{$-0.48$\\[-1pt]{\scriptsize[$-1.03$,\,$+0.55$]}} & \makecell[r]{$+0.14$\\[-1pt]{\scriptsize[$-0.06$,\,$+0.35$]}} \\
advio-02 & reference & 267 & \makecell[r]{$-0.92$\\[-1pt]{\scriptsize[$-1.24$,\,$-0.63$]}} & \makecell[r]{$-0.72$\\[-1pt]{\scriptsize[$-1.33$,\,$-0.24$]}} & \makecell[r]{$+0.10$\\[-1pt]{\scriptsize[$-1.42$,\,$+0.92$]}} & \makecell[r]{$-0.34$\\[-1pt]{\scriptsize[$-0.56$,\,$-0.15$]}} \\
advio-03 & ARKit & 178 & \makecell[r]{$-0.19$\\[-1pt]{\scriptsize[$-0.41$,\,$+0.00$]}} & \makecell[r]{$-0.03$\\[-1pt]{\scriptsize[$-0.66$,\,$+0.52$]}} & \makecell[r]{$+0.24$\\[-1pt]{\scriptsize[$-1.65$,\,$+0.75$]}} & \makecell[r]{$+0.22$\\[-1pt]{\scriptsize[$+0.06$,\,$+0.40$]}} \\
advio-04 & reference & 729 & \makecell[r]{$-0.15$\\[-1pt]{\scriptsize[$-0.31$,\,$+0.06$]}} & \makecell[r]{$-0.28$\\[-1pt]{\scriptsize[$-0.54$,\,$+0.11$]}} & \makecell[r]{$-0.49$\\[-1pt]{\scriptsize[$-0.90$,\,$+0.14$]}} & \makecell[r]{$+1.42$\\[-1pt]{\scriptsize[$+1.28$,\,$+1.53$]}} \\
advio-05 & reference & 301 & \makecell[r]{$+0.42$\\[-1pt]{\scriptsize[$+0.18$,\,$+0.68$]}} & \makecell[r]{$+0.46$\\[-1pt]{\scriptsize[$+0.12$,\,$+0.79$]}} & \makecell[r]{$+0.24$\\[-1pt]{\scriptsize[$-0.15$,\,$+0.75$]}} & \makecell[r]{$+0.42$\\[-1pt]{\scriptsize[$+0.25$,\,$+0.59$]}} \\
advio-20 & ARKit & 133 & \makecell[r]{$-0.34$\\[-1pt]{\scriptsize[$-0.57$,\,$-0.15$]}} & \makecell[r]{$-0.79$\\[-1pt]{\scriptsize[$-1.09$,\,$-0.50$]}} & \makecell[r]{$-1.14$\\[-1pt]{\scriptsize[$-1.62$,\,$-0.77$]}} & \makecell[r]{$-0.02$\\[-1pt]{\scriptsize[$-0.14$,\,$+0.09$]}} \\
advio-21 & ARKit & 132 & \makecell[r]{$+0.17$\\[-1pt]{\scriptsize[$-0.07$,\,$+0.46$]}} & \makecell[r]{$+0.17$\\[-1pt]{\scriptsize[$-0.41$,\,$+0.79$]}} & \makecell[r]{$-0.41$\\[-1pt]{\scriptsize[$-0.81$,\,$-0.02$]}} & \makecell[r]{$+0.18$\\[-1pt]{\scriptsize[$+0.05$,\,$+0.32$]}} \\
\bottomrule
\end{tabular}

\caption{Held-out clips, per clip, as in \cref{tab:perclip_dev}. The protocol
specified the per-clip results of the final forecaster; Tier A $-$ CV was not
specified.}
\label{tab:perclip_heldout}
\end{table}

\begin{table}[ht]
\centering\small
\begin{tabular}{@{}lrrrrrrrr@{}}
\toprule
 & \multicolumn{2}{c}{forecasts scored} & \multicolumn{2}{c}{within 8\,m} & \multicolumn{2}{c}{time with no track} & \multicolumn{2}{c}{NLL $-$ CV @1.2\,s} \\
\cmidrule(lr){2-3}\cmidrule(lr){4-5}\cmidrule(lr){6-7}\cmidrule(lr){8-9}
Clip & fixed & follows & fixed & follows & fixed & follows & fixed & follows \\
\midrule
advio-01 & 124 & 130 & 0\,\% & 19\,\% & 40\,\% & 38\,\% & $-0.31$ & $-0.30$ \\
advio-02 & 276 & 267 & 9\,\% & 10\,\% & 0\,\% & 0\,\% & $-0.82$ & $-0.92$ \\
advio-03 & 175 & 178 & 37\,\% & 40\,\% & 11\,\% & 11\,\% & $-0.12$ & $-0.19$ \\
advio-04 & 28 & 729 & 82\,\% & 98\,\% & 93\,\% & 2\,\% & $+0.09$ & $-0.15$ \\
advio-05 & 299 & 301 & 68\,\% & 68\,\% & 7\,\% & 7\,\% & $+0.43$ & $+0.42$ \\
advio-20 & 134 & 133 & 7\,\% & 6\,\% & 53\,\% & 53\,\% & $-0.37$ & $-0.34$ \\
advio-21 & 133 & 132 & 5\,\% & 4\,\% & 62\,\% & 62\,\% & $0.00$ & $+0.17$ \\
\bottomrule
\end{tabular}

\caption{Held-out clips: the final forecaster with the floor held fixed at the
first frame versus following the phone (pre-registered secondary analysis);
columns as in \cref{tab:floor_dev}.}
\label{tab:floor_heldout}
\end{table}

\begin{figure}[ht]
\centering
\includegraphics[width=\linewidth]{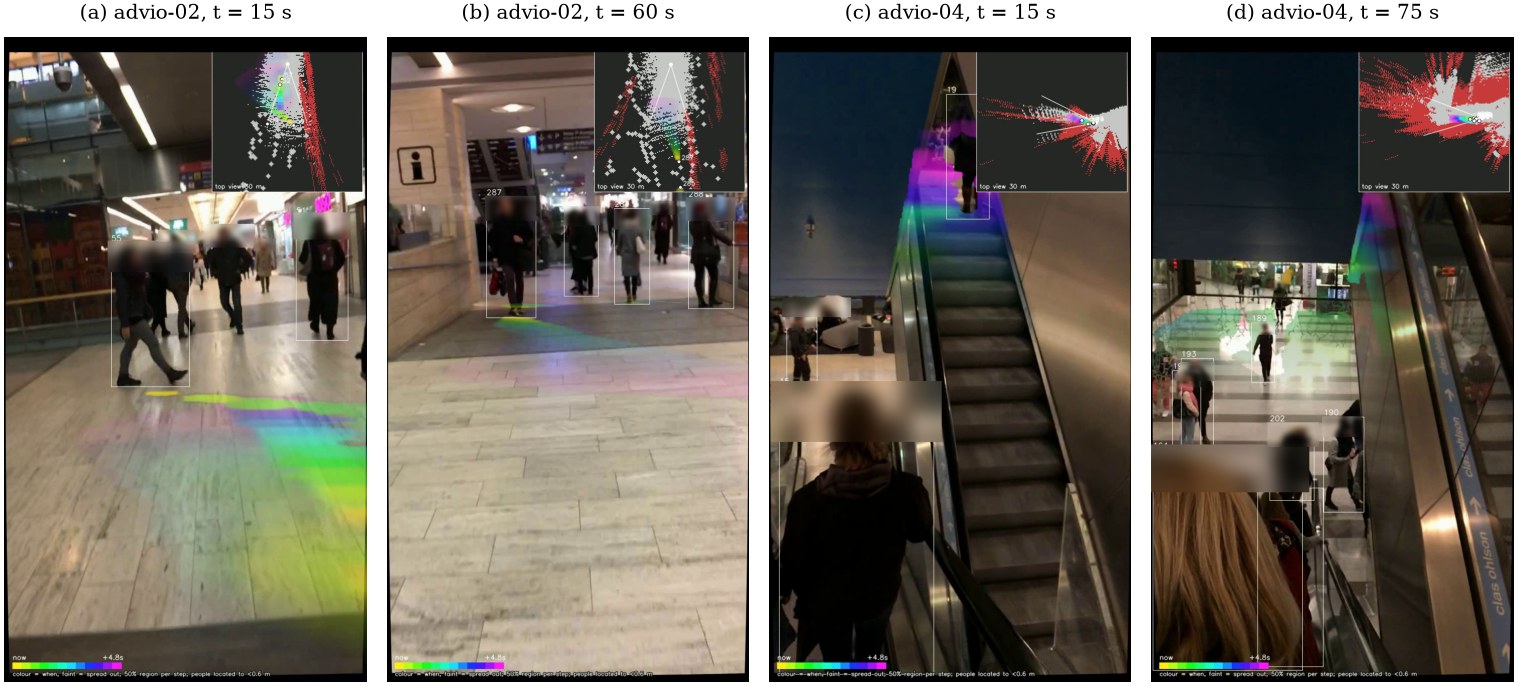}
\caption{Frames \emph{selected by us after the runs} from the final
forecaster's held-out videos, to show typical output with several people and
two failure modes. (a, b) advio-02: forecasts of people walking towards and
past the phone; near the camera the 4.8\,s regions cover much of the floor.
(c) advio-04: a person at the top of an escalator, whose forecast is drawn on
the escalator steps. (d) advio-04: seen from an escalator, people on the floor
below; the assumption of one level floor a camera height below the phone does
not hold here, so their metric positions are wrong even where the overlay,
drawn with the same floor, looks plausible. Colours and anonymisation as in
\cref{fig:frames}.}
\label{fig:frames_selected}
\end{figure}

\begin{table}[ht]
\centering\small\setlength{\tabcolsep}{4pt}
\begin{tabular}{@{}llrrrrrrrrr@{}}
\toprule
 & & & \multicolumn{2}{c}{one width} & \multicolumn{3}{c}{CV mixture} & \multicolumn{3}{c}{final + CV component} \\
\cmidrule(lr){4-5}\cmidrule(lr){6-8}\cmidrule(lr){9-11}
Ahead & Distance & $n$ (dev) & CV $s'$ & blur & $s_1$ & $s_2$ & $a$ & blur & $s$ & $a$ \\
\midrule
1.2\,s & <8 m & 361 & 0.44 & 0.00 & 0.17 & 0.79 & 0.33 & 0.00 & 0.09 & 0.39 \\
 & 8--15 m & 1,069 & 0.46 & 0.00 & 0.12 & 1.06 & 0.29 & 0.00 & 0.00 & 0.41 \\
 & >15 m & 1,705 & 1.02 & 0.00 & 0.00 & 2.17 & 0.35 & 0.00 & 2.59 & 0.03 \\
\addlinespace
2.4\,s & <8 m & 158 & 0.71 & 0.00 & 0.30 & 1.31 & 0.36 & 0.00 & 0.12 & 0.20 \\
 & 8--15 m & 703 & 0.92 & 0.00 & 0.37 & 1.52 & 0.41 & 0.25 & 0.11 & 0.19 \\
 & >15 m & 1,276 & 2.23 & 0.50 & 0.97 & 4.28 & 0.41 & 0.00 & 3.65 & 0.10 \\
\addlinespace
4.8\,s & <8 m & 48 & 1.09 & 0.00 & 0.42 & 1.59 & 0.54 & 0.00 & 0.39 & 0.18 \\
 & 8--15 m & 276 & 1.37 & 0.00 & 0.75 & 3.03 & 0.37 & 0.00 & 0.50 & 0.13 \\
 & >15 m & 673 & 5.71 & 2.00 & 2.62 & 11.77 & 0.45 & 2.00 & 35.29 & 0.02 \\
\bottomrule
\end{tabular}

\caption{Exploratory recalibration (\cref{sec:exp_explore}): values fitted on
the development clips per horizon and distance band ($n$: development
forecasts). One width: the CV baseline's width $s'$ (m, added in quadrature to
the measurement noise) and the extra blur of the final forecaster (m).
Mixtures: the CV baseline as two Gaussians with widths $s_1 < s_2$ (m, each
added in quadrature to the measurement noise) and weight $a$ on the second;
the final forecaster after its blur, mixed with weight $a$ with a Gaussian
around the constant-velocity mean of width $s$ (m, likewise).}
\label{tab:recal_fits}
\end{table}

\begin{figure}[ht]
\centering
\includegraphics[width=\linewidth]{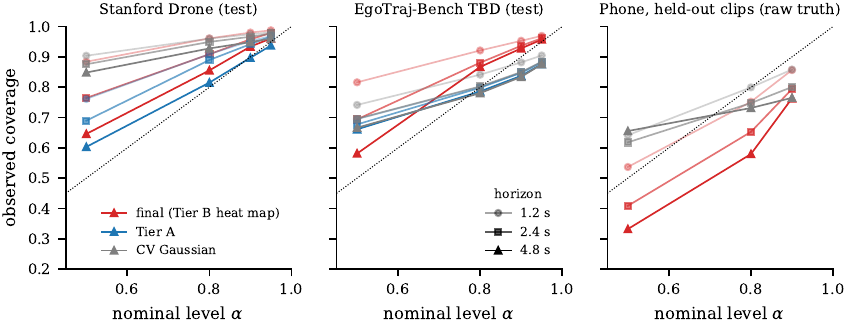}
\caption{Calibration: share of the truths inside the $\alpha$-highest-density
region against $\alpha$, at 1.2, 2.4 and 4.8\,s (lighter to darker). Left and
middle: the benchmark test splits (true positions), the final forecaster,
Tier A and CV, $\alpha = 0.5, 0.8, 0.9, 0.95$. Right: the held-out phone clips
(raw measurements), the final forecaster and CV only, $\alpha = 0.5, 0.8, 0.9$.
Perfect calibration lies on the diagonal; above it, the regions are too wide.}
\label{fig:calibration}
\end{figure}

\section{Pre-registered protocol}
\label{app:protocol}
The protocol below was committed to the project repository (commit
\texttt{\ProtocolCommit}) before any forecast was run on the held-out clips.
It is reproduced verbatim except for formatting.

{\raggedright
{\small\sloppy

Written and committed \textbf{before} any forecast was run on the clips below.
Everything here is fixed; the results will be reported in full whatever they are.

\needspace{4\baselineskip}\paragraph{1. Clips}

ADVIO (Cortés et al., ECCV 2018; Zenodo 10.5281/zenodo.1476931, CC BY-NC 4.0),
sequences \textbf{01, 02, 03, 04, 05 (mall), 20, 21 (outdoor)} -- 7 clips, none of which
was used for any development decision. They were downloaded (md5-checked) and
converted with the unchanged converter (\path{scripts/convert_advio.py}, converter
version 4): the 90 s window with the most detected people, upright undistorted
frames at 30 fps, ARKit poses unless ARKit fails the whole-sequence plausibility
check (then ADVIO's ground-truth poses: clips 02, 04, 05), camera height from the
people in the window (people as the ruler; fallback 1.40 m when implausible: clips
02, 04). What was looked at before this protocol: the converter's console output
(windows, pose choice, camera-height estimates) and its three preview JPEGs per
clip. No tracking, forecasting or scoring was run on them.

Development clips (all decisions so far): advio-06, 11, 12, 22, 23.

\needspace{4\baselineskip}\paragraph{2. Frozen configuration}

\begin{itemize}[leftmargin=*]
\item Code: the commit that adds this file; no change to \texttt{code/},
\path{scripts/eval_phone_clips.py} or \path{scripts/bootstrap_ci.py} until all held-out
runs and their analysis are finished (paper text, tables and figures may change).
\item Pipeline: \texttt{code/world/run\_world.py -{}-source stray -{}-path \textless{}clip\textgreater{} -{}-score \textless{}json\textgreater{}}
on this PC's CPU (Anaconda Python, \path{REALTRAJ_ORT_THREADS=4}), exactly as the
development runs (\texttt{scripts/eval\_phone\_clips.py -{}-threads 4}).
\item Tier B weights: \path{runs/colab_a100_run1/goalnet_selected.pt},
md5 \texttt{\seqsplit{eb0a0832b42bc438881f896e3555b8d9}} (seed 0, chosen on validation).
\item Fusion weights: \path{code/world/params/tierB_fusion.json} (main 1.0, noisy 1.0;
chosen on SDD-val / EgoTraj-val), md5 \texttt{\seqsplit{4b989ab82735ead9c0a4769317a25826}}.
\item Tier A parameters: \path{code/world/params/tierA_params_sdd.json}
(md5 \texttt{\seqsplit{1ad35689c57e219d2863c514b051de0e}}) and \path{tierA_params_phone.json}
(md5 \texttt{\seqsplit{918e8c5b3ef4c744e07d8b70e60504f1}}), fitted on training splits.
\item Detector \texttt{yolov8n.onnx} (md5 \texttt{\seqsplit{12a62296d5ea9b372005ce979e39a356}}); segmentation
\path{models_and_data/scene_seg_logits.onnx} (md5 \texttt{\seqsplit{b5b9bbecc555fcd9da3fd265ce3abeab}}).
\item Floor follows the smoothed camera height, time constant 2 s (default).
\item Segmentation in the forecasting map without LiDAR: \textbf{on} (\texttt{-{}-semantic\_map on}),
decided on the five development clips by the rule in section 5: paired difference
(map on minus off) $-$0.029 / $-$0.015 / $-$0.027 nats at 1.2 / 2.4 / 4.8 s, mean $-$0.024
(\path{runs/paper/dev/map_decision.json}, \path{scripts/decide_map.py}).
\end{itemize}
The final forecaster (``final'') is \texttt{tierB\_net} with that map setting, i.e. the
variant \texttt{tierB\_net\_mapon}.

\needspace{4\baselineskip}\paragraph{3. Runs (each once; a run that crashes for a technical reason is re-run)}

Command: \texttt{scripts/eval\_phone\_clips.py -{}-out runs/paper/heldout -{}-threads 4 -{}-variants \textless{}below\textgreater{}}
with \texttt{-{}-clips advio-01 advio-02 advio-03 advio-04 advio-05 advio-20 advio-21}.

\begin{center}\footnotesize
\begin{tabular}{@{}p{0.42\linewidth}p{0.5\linewidth}@{}}
\toprule
variant & purpose \\
\midrule
\texttt{tierB\_net\_mapon} (final) & primary \\
\texttt{tierB\_net} (the other map setting) & secondary: does the development map decision hold? \\
\texttt{tierA} & secondary: learned heat map vs physics prior \\
\texttt{tierB} (goals + physics) & secondary \\
\path{tierB_net_mapon_fixedfloor} & secondary: the floor fix on fresh clips \\
\path{tierB_net_mapon_h147} (camera height fixed at 1.47 m, the median of the development clips' estimates) & sensitivity: camera-height calibration \\
\bottomrule
\end{tabular}
\end{center}

Videos are saved for the final variant only (\texttt{-{}-videos tierB\_net\_mapon}). The paper's qualitative figure shows the
frame at t = 45 s (the middle) of the ``final'' videos of advio-03, advio-05 and
advio-20, whatever they show.

\needspace{4\baselineskip}\paragraph{4. Analysis}

\begin{itemize}[leftmargin=*]
\item Scores: NLL of the raw measurement nearest each horizon (1.2 / 2.4 / 4.8 s),
forecast and baseline widened by its noise (\texttt{world/scoring.py}, ``raw'' truth).
Also reported: the RTS-smoothed truth, coverage of the 50/80/90 \% regions,
distance bands (\textless{} 8, 8-15, \textgreater{} 15 m), every clip separately.
\item \textbf{Primary comparison:} final minus the constant-velocity Gaussian (CV), raw
truth, all distances, pooled over the 7 clips, at each horizon: mean paired
difference with a 95 \% percentile bootstrap interval (2,000 resamples, seed 0)
that resamples people (clip, track id) -- \texttt{scripts/bootstrap\_ci.py phone}.
\textbf{Hypothesis:} the difference is negative at all three horizons (interval
entirely below 0).
\item Secondary: final minus Tier A, final minus the other map setting (both paired,
same intervals); the floor ablation and the sensitivity run change the
geometry and hence the scored forecasts, so for them each variant is
compared with its own CV (paired within the variant), plus the number of
scored forecasts.
\item No clip or forecast is excluded.
\end{itemize}
\needspace{4\baselineskip}\paragraph{5. Map decision rule (development clips only)}

Pooled over the five development clips, raw truth, all distances: if the mean
over the three horizons of the paired NLL difference (tierB\_net with segmentation
in the map minus tierB\_net without) is negative, the final configuration uses
\texttt{-{}-semantic\_map on}; otherwise the default (\texttt{auto}: segmentation only with LiDAR).
\par}
\par}

\section{Reproducibility}
\label{app:repro}
\paragraph{Software and hardware.} All phone-clip runs and benchmark
evaluations ran on the CPU of a workstation with an Intel Xeon w7-2495X
(24 cores), 63\,GB of RAM and an NVIDIA RTX 4000 Ada (20\,GB), under Windows 11,
with Python 3.12.12, NumPy 2.5.1, SciPy 1.18.1, OpenCV 4.13.0 (whose DNN
module ran the person detector), ONNX Runtime 1.24.4 (segmentation) and
PyTorch 2.11.0 (CPU build; the network). The additional training seeds and
the GPU timings used Python 3.12.14, NumPy 2.5.3, OpenCV 5.0.0, ONNX Runtime
1.26.0 (CUDA; detector and segmentation in the GPU timing) and PyTorch 2.11.0
with CUDA 12.8; seed 0 was trained on Google Colab (A100, PyTorch 2.11.0).
Phone-clip runs used four threads each and several runs shared the CPU, so
their wall-clock speed is not meaningful; speed is measured separately
(\cref{sec:exp_runtime}). Evaluation is deterministic: re-running the final
configuration on all twelve clips reproduced every stored score exactly
(\cref{sec:exp_explore}), as did the final benchmark runs against earlier
runs of the same configuration.

\paragraph{Data.} ADVIO sequences from Zenodo (record 1476931, CC BY-NC 4.0),
md5-checked; SDD TrajNet split and semantic maps as distributed with
Y-Net~\citep{mangalam2021goals}; OpenTraj's SDD scale estimates~\citep{amirian2021opentraj};
EgoTraj-Bench L2-processed data, TBD subset v1.1~\citep{liu2026egotraj}.

\paragraph{Checksums of the frozen artefacts.}
\begin{center}\footnotesize
\begin{tabular}{@{}ll@{}}
\toprule
File & MD5 \\
\midrule
Tier B weights, seed 0 (\texttt{goalnet\_selected.pt}) & \texttt{eb0a0832b42bc438881f896e3555b8d9} \\
Fusion weights (\texttt{tierB\_fusion.json}) & \texttt{4b989ab82735ead9c0a4769317a25826} \\
Tier A parameters, clean-track (SDD) profile & \texttt{1ad35689c57e219d2863c514b051de0e} \\
Tier A parameters, noisy-track (EgoTraj-Bench) profile\footnotemark & \texttt{918e8c5b3ef4c744e07d8b70e60504f1} \\
Person detector (\texttt{yolov8n.onnx}) & \texttt{12a62296d5ea9b372005ce979e39a356} \\
Segmentation (\texttt{scene\_seg\_logits.onnx}) & \texttt{b5b9bbecc555fcd9da3fd265ce3abeab} \\
\bottomrule
\end{tabular}
\end{center}
\footnotetext{Its file is called \texttt{tierA\_params\_phone.json} because it
is used for most people on the phone; it was fitted on EgoTraj-Bench, not on
phone data. The project's working name, RealTraj, also appears in file and
variable names (e.g.\ \texttt{REALTRAJ\_ORT\_THREADS}); it is unrelated to
RealTraj~\citep{fujii2026realtraj}.}

{\sloppy
\paragraph{Commands.} Recordings: \path{scripts/convert_advio.py};
Tier A parameters: \path{scripts/fit_tier_a.py},
\path{scripts/fit_phone_profile.py} and \path{scripts/fit_mixture.py};
network: \path{code/world/train_goalnet.py} (with \texttt{-{}-legacy\_mirror}
to reproduce the reported networks, \cref{app:training}); fusion weight:
\path{scripts/select_tierb_fusion.py}; map decision:
\path{scripts/decide_map.py}; phone clips: \path{scripts/eval_phone_clips.py}
(variants as in \cref{app:protocol}); benchmarks:
\path{scripts/paper_benchmarks.py}; paired bootstrap intervals:
\path{scripts/bootstrap_ci.py}; timing: \path{scripts/paper_latency.py};
synthetic floor test: \path{scripts/paper_synthetic_floor.py}; tables and
figures: \path{paper/make_tables.py} and \path{paper/make_figures.py}.
\par}

\paragraph{Test-set usage.} The SDD test split was evaluated \SddTestUses{}
times and the EgoTraj-Bench test split \EgoTestUses{} times over the whole
project, counting the final runs and the two additional seeds; the list is in
\path{paper/make_tables.py}. No parameter was chosen on a test split.

\paragraph{Licences.} ADVIO and EgoTraj-Bench are released under CC BY-NC 4.0
and the Stanford Drone Dataset under CC BY-NC-SA 3.0 (non-commercial use);
the YOLOv8 detector is AGPL-3.0. We use all of them for non-commercial
research only.

\section{Use of AI assistance}
\label{app:ai}
The software, the
experiments and a first draft of this manuscript were produced with the help
of an AI assistant (Claude, Anthropic) operated by the author, who directed
the work, checked the results and is responsible for the content.

\end{document}